%% file: main.tex
\documentclass[10pt,twocolumn,letterpaper]{article}

\usepackage[pagenumbers]{cvpr}      

\input{preamble}

\definecolor{cvprblue}{rgb}{0.21,0.49,0.74}
\usepackage[pagebackref,breaklinks,colorlinks,allcolors=cvprblue]{hyperref}
\usepackage{makecell}
\usepackage{adjustbox}
\newcommand{\akira}[0]{AKiRa\xspace}

\def\paperID{1350} 
\def\confName{CVPR}
\def\confYear{2025}

\title{\akira: Augmentation Kit on Rays for optical video generation}

\author{
Xi Wang\textsuperscript{1}, Robin Courant\textsuperscript{1}, Marc Christie\textsuperscript{2}, Vicky Kalogeiton\textsuperscript{1}\\
\textsuperscript{1}LIX, École Polytechnique, IP Paris \quad \textsuperscript{2}Univ Rennes, IRISA, Inria, CNRS\\
}

\usepackage{cuted}
\usepackage{capt-of}

\begin{document}
\maketitle

\begin{strip}
\centering
\includegraphics[width=\textwidth]{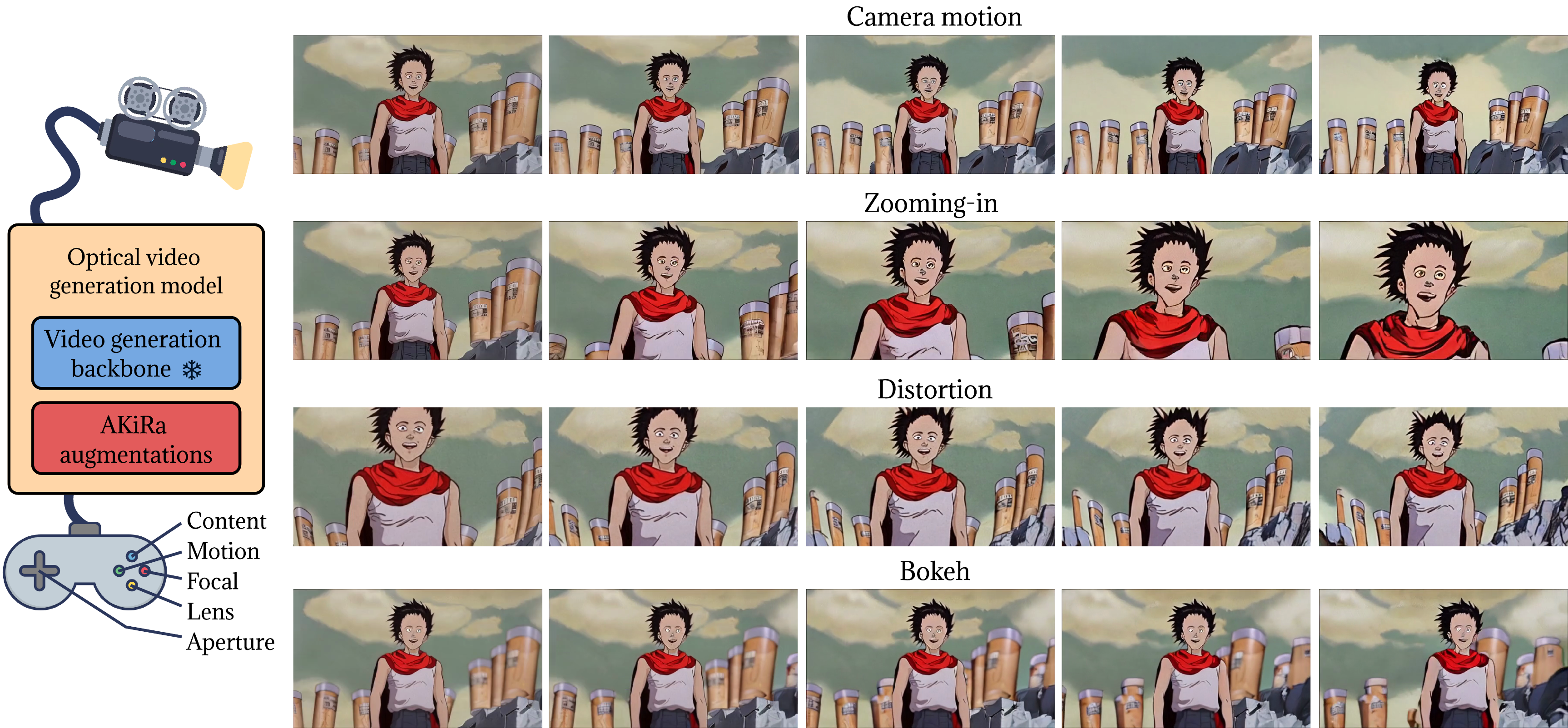}
\captionof{figure}{While current state-of-the-art video generation approaches offer limited control to users on camera motion, we propose a dedicated data augmentation framework —\akira— to train an optical video generation model that provides users with a panel of controls on camera motions (top row), camera focal length (second row), lens distortion (third row), or bokeh (camera aperture an in/out of focus regions in bottom row). See more in our \href{https://www.lix.polytechnique.fr/vista/projects/2024_akira_wang}{project page}.}
\label{fig:feature-graphic}
\end{strip}

\input{sec/0.abstract}    

\section{Introduction}
\label{sec:intro}
\input{sec/1.intro}

\section{Related work}
\label{sec:related-work}
\input{sec/2.related-work}

\section{Method}
\label{sec:method}
\input{sec/3.method}

\section{Experiments}
\label{sec:experiments}
\input{sec/4.experiments}

\section{Conclusion}
\label{sec:conclusion}
\input{sec/5.conclusion}

\section*{Acknowledgments}
\label{sec:acknowledgements}
\input{sec/acknowledgements}

{
\bibliographystyle{splncs04}
\bibliography{short-strings,references}
}


\newpage
\input{supmat}

\end{document}

%% file: preamble.tex
\usepackage{algorithm}
\usepackage{algpseudocode}
\usepackage{array}
\usepackage{booktabs}
\usepackage{multirow}
\usepackage{dsfont}
\usepackage{color}
\usepackage{colortbl}

\definecolor{tabfirst}{rgb}{1, 0.7, 0.7} 
\definecolor{tabsecond}{rgb}{1, 0.85, 0.7} 
\definecolor{tabthird}{rgb}{1, 1, 0.7} 

%% file: sec/0.abstract.tex
\begin{abstract}
Recent advances in text-conditioned video diffusion have greatly improved video quality.
However, these methods offer limited or sometimes no control to users on camera aspects, including dynamic camera motion, zoom, distorted lens and focus shifts. 
These motion and optical aspects are crucial for adding controllability and cinematic elements to generation frameworks, ultimately resulting in visual content that draws focus, enhances mood, and guides emotions according to filmmakers' controls. In this paper, we aim to close the gap between controllable video generation and camera optics.
To achieve this, we propose \akira (Augmentation Kit on Rays), a novel augmentation framework that builds and trains a camera adapter with a complex camera model over an existing video generation backbone.
It enables fine-tuned control over camera motion as well as complex optical parameters (focal length, distortion, aperture) to achieve cinematic effects such as zoom, fisheye effect, and bokeh. 
Extensive experiments demonstrate \akira's effectiveness in combining and composing camera optics while outperforming all state-of-the-art methods. This work sets a new landmark in \textbf{controlled} and \textbf{optically enhanced} video generation, paving the way for future optical video generation methods. 
\end{abstract}

%% file: sec/1.intro.tex
Creating high-quality video sequences has always been a complex interplay between the content itself and the means, the camera, by which it is portrayed. A powerful narrative or stunning visual concept can only reach its full potential when the camera effectively translates it into moving images, capturing nuances through cinematic effects such as dynamic camera motions, zooms, distorted lenses, or intentional focus shifts. Such techniques add texture and depth, drawing the viewer’s eye to specific details, enhancing mood, and subtly guiding emotions. When mastered, they allow filmmakers to transform raw content into a rich, immersive experience that feels artful and purposeful.

Recently, video generation have gained notable popularity in the community~\cite{guo2023animatediff,blattmann2023svd}. Researchers have focused on various aspects, such as improving generation quality~\cite{opensora,davtyan2023river,polyak2024movie}, increasing resolution~\cite{ho2022imagen,singer2022make} or enhancing the efficiency~\cite{blattmann2023align,an2023latent,wang2023lavie}. 
Despite the remarkable results of modern work in most aforementioned aspects, they often overlook the cinematic techniques, crucial for achieving realistic and immersive storytelling. 
To tackle this, some recent approaches~\cite{wang2024motionctrl,he2024cameractrl} provide some control over camera motion. Specifically, given raw camera poses and text, these methods generate video following the specified camera trajectory.
However, all these approaches simplify the camera model to its \emph{motion} alone and neglect crucial \textbf{optical effects} like \emph{focal length (for zoom), lens distortion (fisheye effect), aperture and focus point (for focus shifts)} for three reasons. 
First, most methods do not use any underlying camera model~\cite{opensora,davtyan2023river,polyak2024movie}, and instead, they treat videos as a sequence of pixels; hence, despite excessive training, the underlying representations still lack visual coherency. 
Second, some recent works~\cite{wang2024motionctrl,he2024cameractrl} use simplistic camera models, treating videos as a sequence of posed cameras without lens effects; hence, lack control. 
Third, there is no adequate training data with optical effects or optical parameters meaning that methods cannot use the data to train for optical camera effects directly. 
Accounting for such optical parameters attached to the camera is essential, and their absence limits the generation of optically coherent video content and reduces the potential for cinematic quality. 

To address the above limitations, we introduce the concept of \emph{optical} video generation models — video generation models which are optically coherent and over which users can control the motion of the camera as well as its optical parameters such as focal length, lens distortion, aperture and focus point. 
Such a model requires the integration of optical effects into the generation pipeline so that camera and optical parameters are directly leveraged by the model.

To achieve this, we first propose to rely on a Plücker coordinates to represent an image as a collection of camera rays~\cite{plucker1828analytisch,zhang2024cameras} to ensure coherency. We extend the diffusion-based video generation model of~\cite{he2024cameractrl} by proposing an optically-enhanced camera representation with lens distortion and a dedicated aperture map to encode in- and out-of-focus regions in the screen.  
Second, we propose a dedicated framework, named AKiRa (Augmentation Kit on Rays), which exploits these extended camera ray representations to augment the training dataset, by simulating optical effects on input frames paired with corresponding optical parameters. 
With this data augmentation method, we then train a camera adapter on top of a pre-trained and frozen video generation backbone. 
As a result, we provide the first \emph{optical} video generation framework capable of controlling optical camera parameters in addition to camera trajectories, allowing the creation of complex cinematographic effects. The work contributes to closing the gap between the level of control offered to designers in computer graphics worlds and the visual quality of video generative frameworks. 
We evaluate the generalization of our method across various video generation backbones and on a comprehensive benchmark with robust metrics. 
We show that our method disentangles camera parameters, specifically separating zoom from translational motion—an achievement not possible with other approaches, where these parameters remain intertwined. Finally, AKiRa outperforms state-of-the-art methods —MotionCtrl~\cite{wang2024motionctrl} and CameraCtrl~\cite{he2024cameractrl}. 

Our contributions are:
 (1) the \textbf{first \emph{optical} video generation framework}, which offers the ability to control both camera motions and optics, enabling the generation of videos with complex optical effects (e.g. zoom, fisheye, focus shifts), (2) the design of a \textbf{camera model representation including optical parameters} expressed in a Plücker map, extended with an aperture map to model in- and out-of-focus effect; (3) a \textbf{joint camera-frame Augmentation Kit on Rays (\akira)} modelling optical effects to enable the training of more controllable video generation models.  

%% file: sec/2.related-work.tex
Our work contributes to the field of \emph{controllable video generation}, which can be approached in two ways: in the first, the generation is guided by text, while in the second way, the generation is guided by camera and text.
Another related area is virtual cinematography, which does not generate videos but instead focuses on entities within an existing environment, e.g. camera angles, lighting, and composition. 

\noindent \textbf{Text-to-video generation (T2V).} 
The first video diffusion model is introduced by~\cite{ho2022video}, building on the success of image diffusion~\cite{sohl2015deep,song2020improved,ho2020denoising,dhariwal2021diffusion}. Later, Imagen-Video~\cite{ho2022imagen} and Make-A-Video~\cite{singer2022make} introduced cascaded pixel-based diffusion models for high-definition video generation.
To reduce training costs, some works~\cite{blattmann2023align,an2023latent,wang2023lavie} perform diffusion in latent space. Others~\cite{guo2023animatediff,wang2023modelscope,blattmann2023svd,chen2023videocrafter1} fine-tune temporal adapters on 2D layers of pre-trained text-to-image models~\cite{rombach2022high} using video datasets~\cite{bain2021webvid}.
Recently, \cite{opensora,menapace2024snap,ma2024latte} explored transformer backbones (DiT) for scalability, while RIVER~\cite{davtyan2023river} and MovieGen~\cite{polyak2024movie} moved to flow matching, achieving state-of-the-art performance.
These methods rely \emph{solely} on text guidance, which is often sufficient for image generation. However, video generation requires additional complexity, incorporating temporal dynamics and camera behavior (motion and optics). To address this, we focus on enhancing camera controllability for video generation.

\noindent \textbf{Camera-based T2V.} 
First approaches of T2V \cite{guo2023animatediff,blattmann2023svd} fine-tune the video generation model using LoRA~\cite{hu2022lora} to achieve categorical camera motion. 
Drawing inspiration from the conditional generation problem with auxiliary modules~\cite{dufour2022scam,zhang2023adding,gauthier2014conditional,van2016conditional,ye2023ip}, some works~\cite{wang2024videocomposer,hu2024motionmaster} condition models on motion maps to manage camera motion, while~\cite{wu2024motionbooth,yang2024direct} are restricted to predefined motion directions.
Recent works~\cite{wang2024motionctrl,he2024cameractrl,xu2024camco,bahmani2024vd3d,zheng2024cami2v,cheong2024boosting} directly integrate precise camera pose parameters into video generation:  
MotionCtrl~\cite{wang2024motionctrl} initiated by feeding raw camera poses into the model with a trainable adapter; CameraCtrl~\cite{he2024cameractrl} expanded this work with Pl\"{u}cker coordinates, effectively linking camera intrinsic and extrinsic parameters to image content through ray-based modelling.
However, most research~\cite{xu2024camco,he2024cameractrl,wang2024motionctrl} focuses solely on camera motion, neglecting factors like focal length, distortion, aperture and focus point.
Ignoring focal length, for example, causes current methods to confuse zoom effects with forward and backward translation, resulting in geometric differences~\cite{hartley2003multiple}. 
Moreover, most work assumes a distortion-free camera model, limiting generation capacity, as lens distortions (e.g., fisheye effects) are common in artistic videos~\cite{bertalmio2014image}. Finally, the aperture and focus point are also neglected, despite being powerful effects for guiding image focus. To address these, we propose an \emph{optical video generation model} that incorporates a complex camera model into a video generation backbone and enables control of camera motion, focal length, distortion, aperture and focus point.

\noindent \textbf{Virtual cinematography.} Recent works explore methods for integrating cinematic and aesthetics into content generation, via camera control.
For instance, \cite{jiang2020example,jiang21cam} synthesize camera trajectories with cinematic styles for 3D animations using reference clips.
Cinematic transfer adapts cinematic features from reference clips to new scenes.
JAWS~\cite{wang2023jaws} pioneered this by optimizing camera trajectories directly within a neural radiation field~\cite{mildenhall2020nerf} to match visual cinematic features from a reference clip; follow-up works~\cite{jiang2024cinematic,chen2024dreamcinema,wang2024humanvid} further refined this, notably adding character re-targetting.
Other recent works~\cite{jiang2024ccd,courant2024exceptional} focus on camera trajectory generation using diffusion models conditioned on text prompts and character motion. 
In~\cite{courant2024exceptional} they introduce a dataset of camera trajectories with text extracted from movies, grounding generated trajectories in cinematic features.
Yet, these approaches require existing content (e.g. 3D environments~\cite{jiang2020example,jiang21cam}, clips~\cite{wang2023jaws}), thus limiting creative possibilities. 
This may prevent creators from generating new content, limiting stylistic exploration by encouraging replicating cinematic features from references.
Instead, we propose a method that enables cinematic control directly within video generation, removing the need for pre-existing content and expanding users' creative freedom.

\begin{figure}[t]
    \centering
    \includegraphics[width=\linewidth]{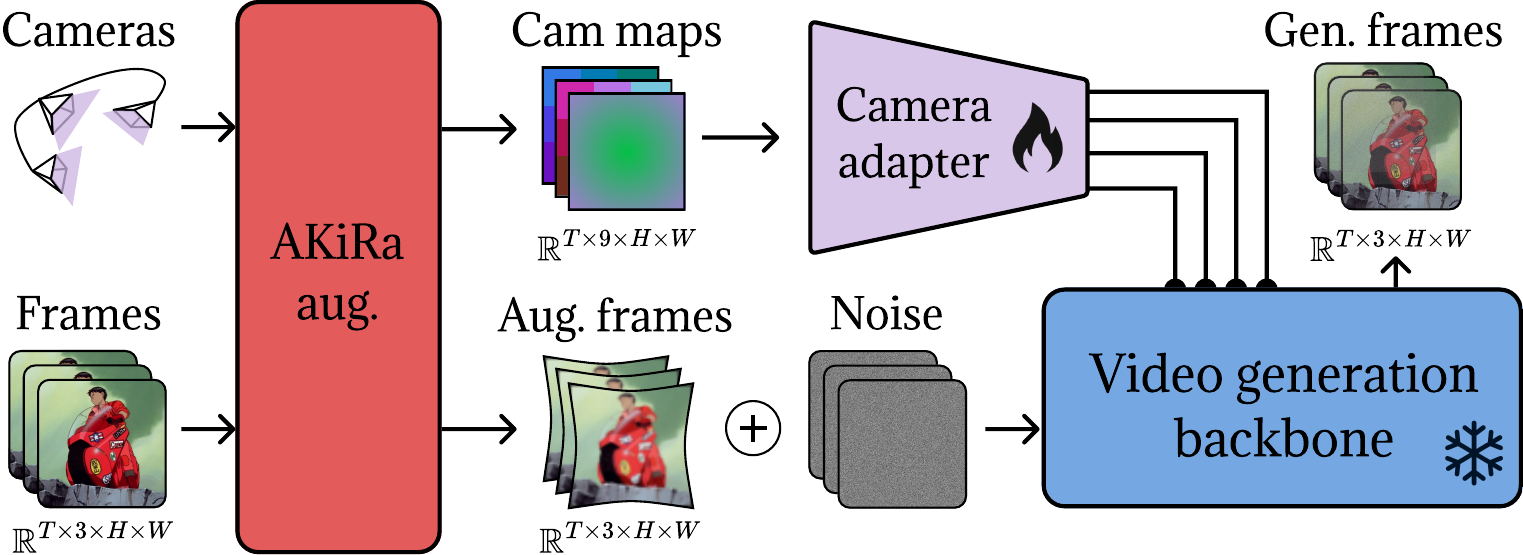}
    \caption{\textbf{Overview of \akira training.} The camera adapter is trained by jointly augmenting camera data and frames using \akira augmentations. The adapter processes multiple camera parameters—motion, focal length, distortion, aperture, and focus point. This adapter is integrated into a pre-trained, frozen backbone, resulting in an optical video generation model.}
    
    \label{fig:overview-akira-train}
\end{figure}

\input{fig/augmentations}

%% file: fig/augmentations.tex
\begin{figure*}[htbp]
    \centering
    \begin{subfigure}{0.24\textwidth}
        \centering
        \includegraphics[width=\linewidth]{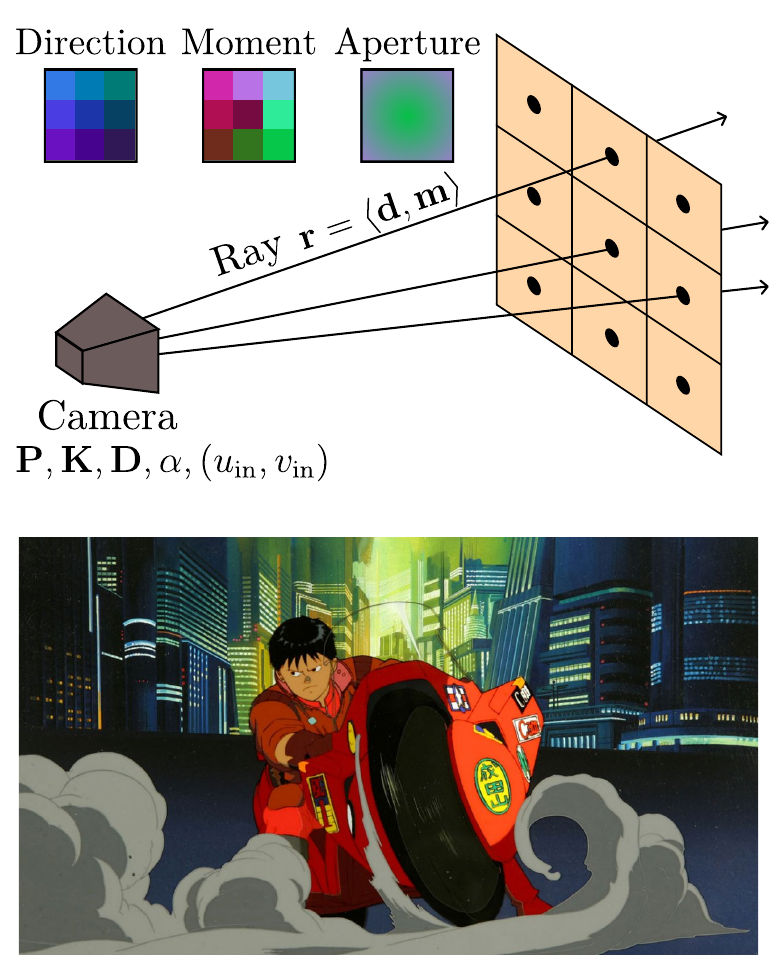}
        \caption{Base}
        \label{fig:base}
    \end{subfigure}
    \hfill
    \begin{subfigure}{0.24\textwidth}
        \centering
        \includegraphics[width=\linewidth]{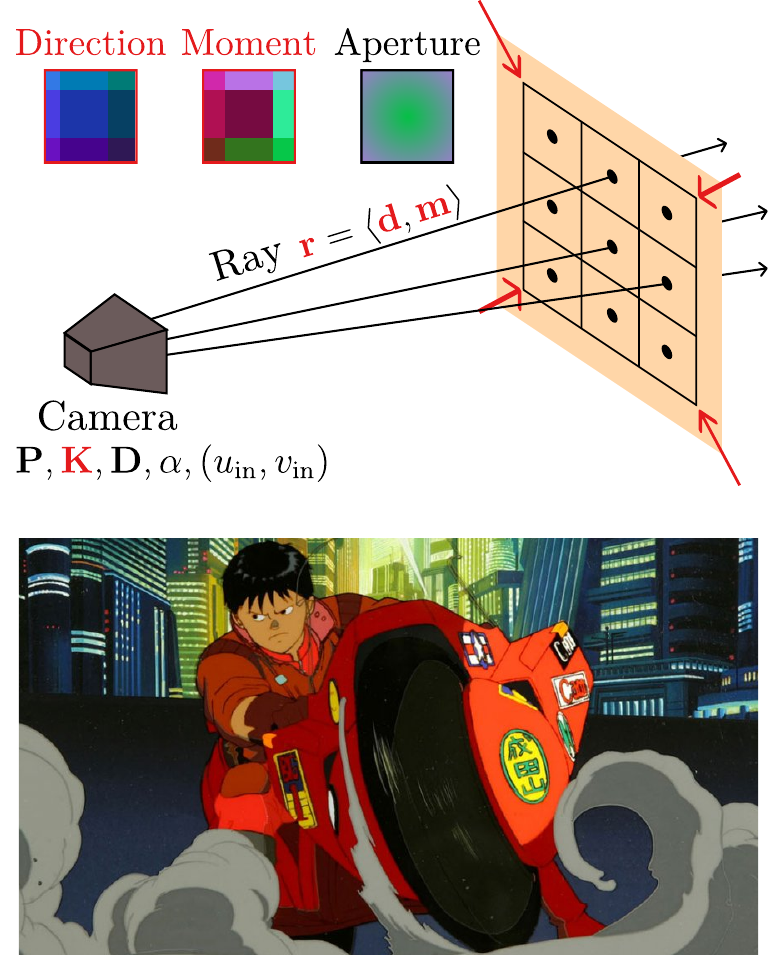}
        \caption{Focal length - Zoom}
        \label{fig:zoom}
    \end{subfigure}
    \hfill
    \begin{subfigure}{0.24\textwidth}
        \centering
        \includegraphics[width=\linewidth]{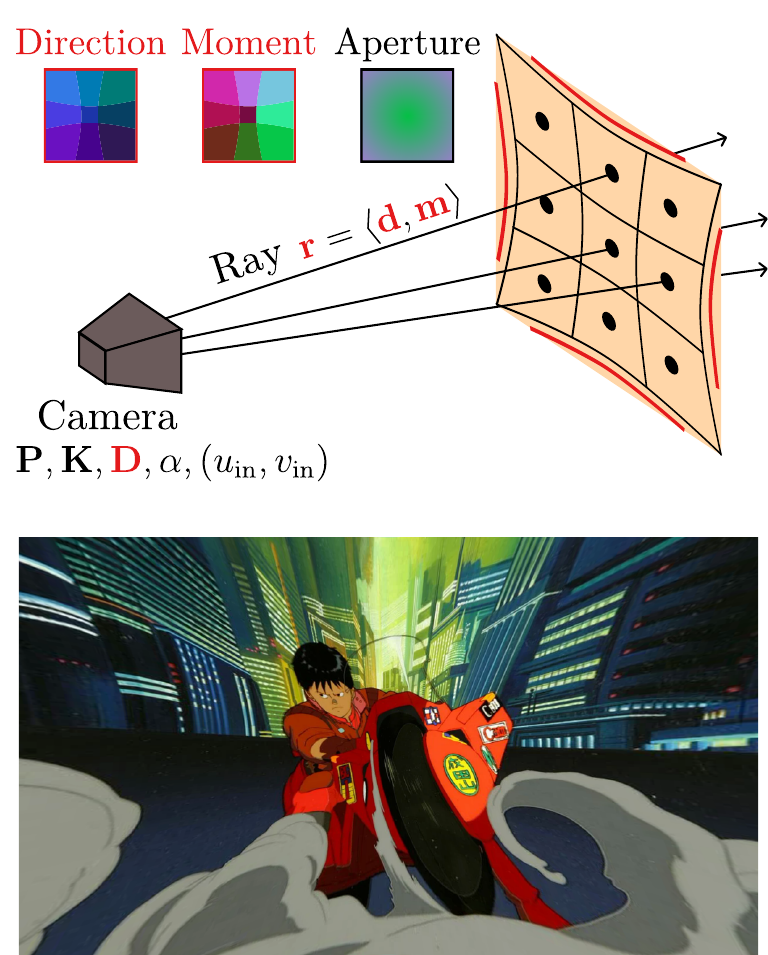}
        \caption{Lens - Distortion}
        \label{fig:distortion}
    \end{subfigure}
    \hfill
    \begin{subfigure}{0.24\textwidth}
        \centering
        \includegraphics[width=\linewidth]{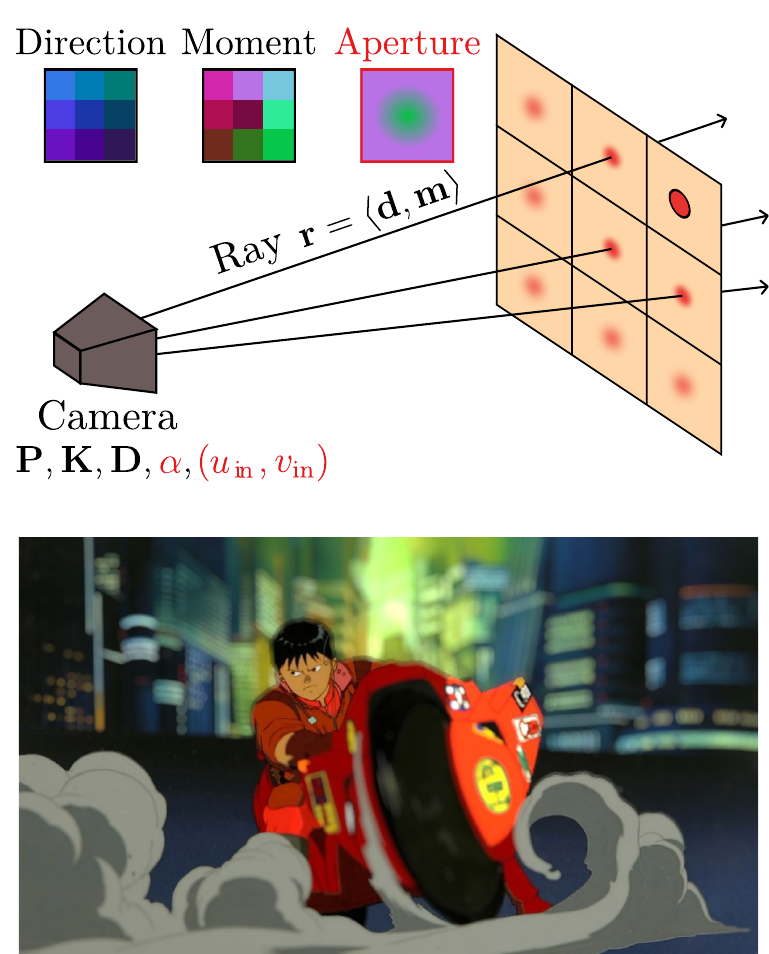}
        \caption{Aperture - Bokeh}
        \label{fig:bokeh}
    \end{subfigure}
    \caption{\textbf{Optical effect overview}. Visualization of various optical effects proposed in our system —zoom, distortion, and bokeh—and their impacts on both the camera parameters (top row) and visual output (bottom row). In addition, as with state-of-art techniques, we enable the control of the camera motion (not displayed here).}
    \label{fig:overview-akira}
\end{figure*}

%% file: sec/3.method.tex
In this section, we present our approach for training an \emph{optical} video generation model, enabling users to manipulate the camera's motion as well as optics to produce cinematic effects such as zoom, distortion, and bokeh. 

\noindent \textbf{Overview.} 
To train an optical video generation model capable of controlling both camera motion and optics, we require training data with associated camera parameters. 
Here, \emph{optics} refer to the camera's optical characteristics, including focal length, lens distortion, aperture and focus point, which together shape how the camera captures field-of-view (zoom), lens characteristics (distortion), and light exposure (in- or out-of-focus). 
Although some recent datasets pair videos with camera trajectories \cite{zhou2018realestate,courant2024exceptional}, to our knowledge, there are no datasets that include videos with rich varying optical information. 
Hence, we propose a set of data augmentations based on a complex camera model that parameterizes \emph{optics} 
(Section~\ref{sub:camera-model}). To better disentangle each parameter, we design a highly expressive representation for our model (Section~\ref{sub:camera-representation}). Finally, we describe the augmentations to generate videos paired with optical parameters (Section~\ref{sub:augmentation}).
As illustrated in Figure~\ref{fig:overview-akira-train}, with these augmentations, we extend prior works~\cite{wang2024motionctrl,he2024cameractrl}, by training a camera adapter that controls camera motion and \emph{optics} 
on top of a pre-trained, frozen video generation backbone.

\subsection{Camera model}
\label{sub:camera-model}

To train an \emph{optical} video generation model, we need a camera model to represent the camera parameters. 
Here we propose a camera model by relying on and extending the pinhole camera model to represent not only camera motion and focal length (pinhole camera model) but also the lens distortion 
and the aperture and focus point (distorted pinhole camera model with aperture), as described below.

\noindent \textbf{Pinhole camera model.} We build our camera 
model upon the standard pinhole camera model~\cite{zisserman2004mvg}, which includes extrinsic and intrinsic camera parameters. 
The \emph{extrinsic parameters} $\mathbf{P} \in \mathbb{R}^{3 \times 4}$  describe the position and orientation of the camera. They are composed of a rotation matrix $\mathbf{R} \in SO(3)$ and translation vector $\mathbf{t} \in \mathbb{R}^{3}$, giving $\mathbf{P} = [\mathbf{R} | \mathbf{t}] \in SE(3)$.
The \emph{intrinsic parameters} $\mathbf{K} \in \mathbb{R}^{3 \times 3}$ include the principal point $(c_x, c_y)$ and the focal length $f \in \mathbb{R}^+$. Specifically, $f$ controls the \emph{zoom-in} (high $f$) and \emph{zoom-out} (low $f$) capabilities of the camera (different from forward or backward translation), as shown in Figure~\ref{fig:zoom}.
Using this pinhole model, a 3D world point $\mathbf{X} \in \mathbb{R}^3$ is projected to 2D pixel coordinates $(u, v) \in \mathbb{R}^2$ as follows:

\begin{equation}
\begin{bmatrix} u \\ v \\ 1 \end{bmatrix} =
\underbrace{
\begin{bmatrix}
f_x & 0 & c_x \\
0 & f_y & c_y \\
0 & 0 & 1
\end{bmatrix}
}_{\text{Intrinsic matrix } \mathbf{K}}
\underbrace{
\begin{bmatrix}
\mathbf{R} & \mathbf{t} \\
0 & 1
\end{bmatrix}
}_{\text{Extrinsic matrix } \mathbf{P}}
\begin{bmatrix}
X \\
1
\end{bmatrix}.
\label{eq:intrinsic_param}
\end{equation}

\noindent \textbf{Distorted pinhole camera model.} The pinhole model does not inherently account for lens distortion, as it assumes a lens-free setup. 
To represent more realistic cameras, we extend this model 
by adding radial distortion parameters $\mathbf{D} \in \mathbb{R}^3$~\cite{slama1980manual} (e.g., the well-known \textit{fisheye lens} with ultra-wide-angle distortion as shown in Figure~\ref{fig:distortion}). 
These parameters adjust the pixel coordinates $(u, v)$ radially to $(u_\mathbf{D}, v_\mathbf{D}) \in \mathbb{R}^2$ with $r = \sqrt{(u - c_x)^2 + (v - c_y)^2}$, follow:
\begin{equation}
    \begin{bmatrix} u_\mathbf{D} \\ v_\mathbf{D} \end{bmatrix} = \begin{bmatrix} u \\ v \end{bmatrix} \left( \mathbf{1}_3 + \mathbf{D} \begin{bmatrix} r^2, r^4, r^6 \end{bmatrix} \right)
    \ ,
\label{eq:distrotion}
\end{equation}

\noindent \textbf{Distorted pinhole camera model with aperture.} 
The standard pinhole model, with infinitely small aperture, does not capture depth-of-field effects.
To simulate the \emph{bokeh effect}—the appearance of in-focus and out-of-focus regions, as illustrated in Figure~\ref{fig:bokeh}—we introduce an aperture parameter $\alpha \in \mathbb{R}$. This parameter, along with focus point $(u_{\text{in}}, v_{\text{in}}) \in \mathbb{R}^2$, controls bokeh intensity and location.

\input{fig/quals}

\subsection{Camera model representation}
\label{sub:camera-representation}

\noindent \textbf{Ray-based camera model representation.} To train our \emph{optical} video generation model, we require an effective camera representation that connects the optical properties of the camera to the generated visual content. 
For this, we map the geometric camera model $(\mathbf{P}, \mathbf{K}, \mathbf{D})$ to screen pixels $(u, v) \in \mathbb{R}^2$ (or patches in practice) using a ray representation $\mathbf{r}$, where each pixel is associated with a ray (a line) passing through the camera's centre $\mathbf{O} \in \mathbb{R}^3$.

\noindent \textbf{Plücker map.} We adopt the Plücker coordinates~\cite{plucker1828analytisch} to represent our camera model, as in~\cite{zhang2024cameras,he2024cameractrl}. A ray $\mathbf{r} = \langle \mathbf{d}, \mathbf{m} \rangle \in \mathbb{R}^6$ is represented by its direction $\mathbf{d}$ and its moment $\mathbf{m}$ about any point $\mathbf{p}$ on the ray, such that $\mathbf{m} = \mathbf{p} \times \mathbf{d}$.
The direction $\mathbf{d}$ is computed by reprojecting the pixel coordinates $(u, v)$ with camera parameters $\mathbf{P}$ and $\mathbf{K}$, and the moment $\mathbf{m}$ is calculated by taking the camera centre $\mathbf{O}$ as the point $\mathbf{p}$ since all rays pass through $\mathbf{O}$:
\begin{equation}
    \mathbf{d} = \mathbf{R}^\top \mathbf{K}^{-1} 
    \begin{bmatrix} u_\mathbf{D}, v_\mathbf{D}, 1 \end{bmatrix}^\top, \quad \mathbf{m} = (-\mathbf{R}^\top\mathbf{t}) \times \mathbf{d} \ .
\end{equation}
Plücker coordinates $\mathbf{m}$ and $\mathbf{d}$ encode both information about the focal length $\mathbf{K}$ and lens distortion $\mathbf{D}$, as these parameters are inherently linked to ray orientations, as shown in Figures~\ref{fig:zoom} and~\ref{fig:distortion}. 
For each frame, we derive direction and moment maps with the same dimensions as the frame, associating each pixel with a specific moment and direction.

\noindent \textbf{Aperture map.} The camera’s aperture parameter $\alpha$ is not captured by Plücker coordinates, as it is not directly ray-related in the Pinhole model, as shown in Figure~\ref{fig:bokeh}.  
To address this, we introduce an aperture map with the same structure and dimensions as the direction and moment maps, assigning an aperture parameter to each pixel in the frame. 
To achieve this, we first define the coordinates of the focus point $(u_\text{in}, v_\text{in})$ representing the sharpest point in the frame. We then define the per-pixel aperture map as $\mathbf{a} \in \mathbb{R}^3$ for any point $(u, v)$ on the frame as:

\begin{equation}
    \mathbf{a} = \begin{bmatrix} (u - u_\text{in}), (v - v_\text{in}), \Vert (u, v) - (u_\text{in}, v_\text{in})\Vert^{\frac{1}{\sigma(\alpha)}} \end{bmatrix}^\top \ .
\end{equation}

\noindent
where $\sigma$ is the sigmoid function. We will discuss the aperture map in the supplementary materials.

Finally, each frame encodes camera information — direction, moment, and aperture $(\mathbf{d}, \mathbf{m}, \mathbf{a})$ — into a 9-dimensional camera map $\mathbb{R}^{9 \times H \times W}$ matching the video frame dimensions.

\subsection{\akira: Augmentation Kit on Rays}
\label{sub:augmentation}

To augment and disentangle optical features in our extended Plücker camera model, we propose \akira, an Augmentation Kit on Rays. It contains augmentation techniques for both video frames and corresponding optical parameters:

\noindent \textbf{Zooming - focal length.} 
For the zooming effect, we augment the focal length $f$ in Equation~\ref{eq:intrinsic_param} using a \textbf{zooming factor} $s \in \mathbb{R}^+$: $s > 1$ represents a zoom-in effect, and $s < 1$ represents zoom-out, both proportional to $s$. 

In image space, changing the focal length can be simulated by a center cropping and resizing the image back to its original resolution. This transformation modifies the image coordinates to $(u_\text{fl}, v_\text{fl})$ after the focal length change:
\begin{equation}
u_\text{fl} = s \left( u - c_x \right) + c_x, \quad v_\text{fl} = s \left( v - c_y \right) + c_y,
\label{eq:focal-change-image}
\end{equation}

\noindent
where ($c_x$, $c_y$) is the principal point of Equation~\ref{eq:intrinsic_param}. With the Plücker map aligned to image pixels, its augmentation can be performed using Equation~\ref{eq:focal-change-image} (see Figure~\ref{fig:zoom}).

\begin{figure}[t]
    \centering
    \includegraphics[width=\linewidth]{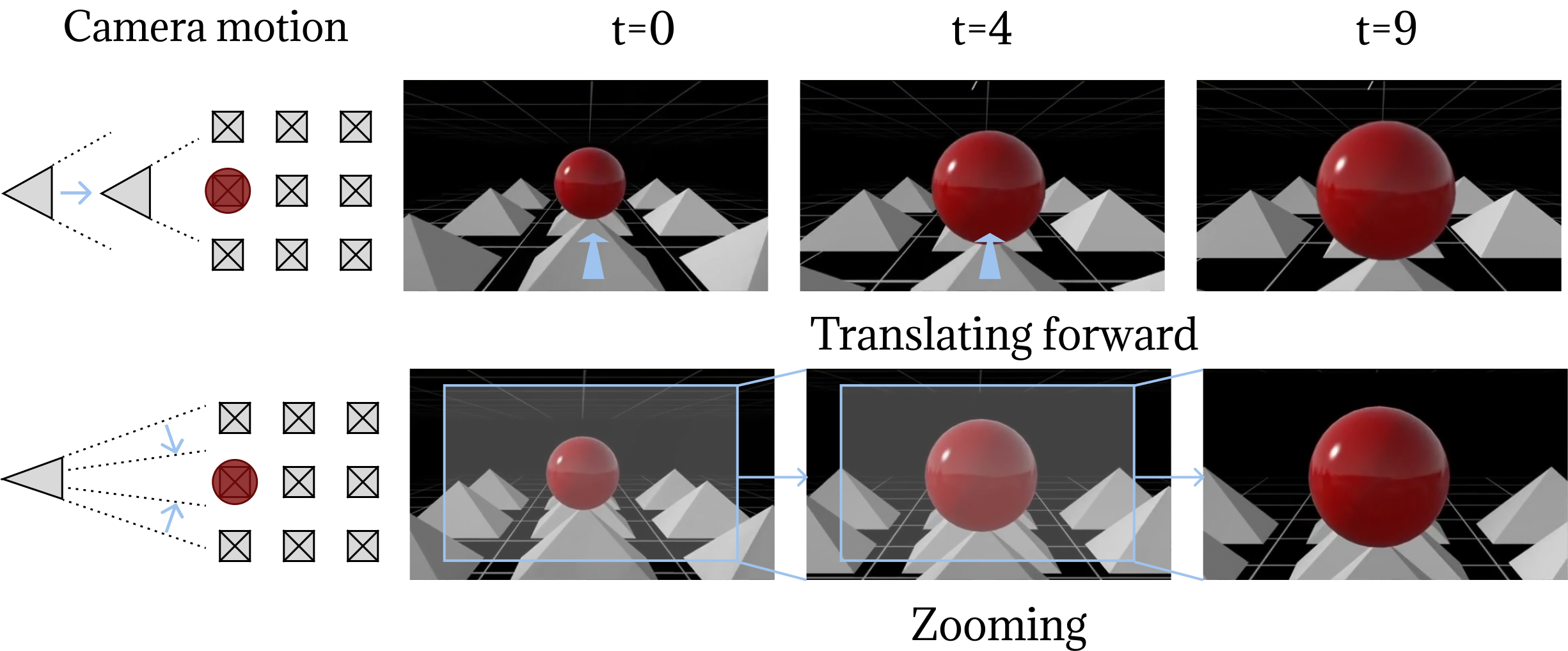}
    \caption{\textbf{Difference between zoom and push forward.} Zooming (change of focal length) is similar to image cropping and resizing while pushing forward changes the perspective of the scene.}
    \label{fig:zoom_vs_push}
\end{figure}

Focal length changes (zoom) are often mistaken for forward/backward movement. While zoom affects only the cropping area, translation induces \textbf{perspective changes} (see Figure~\ref{fig:zoom_vs_push}). 
Such ambiguities can hinder the accurate interpretation of focal length changes and translational motion during training. Thus, augmenting focal length is essential to \textbf{disambiguate} these effects, reducing confusion between zooming and movement, and enabling richer optical compositionality: e.g. moving right while zooming in or moving forward while zooming out (commonly referred to as a dolly zoom), which are popular in modern cinematography.

\noindent \textbf{Distortion - lens.}
Augmentation on distortion involves modifying the radial distortion coefficients $\mathbf{D}$as defined in Equation~\ref{eq:distrotion}. In alignment with focal-length augmentation, this transformation is applied simultaneously to both the image and Plücker coordinates, as shown in Figure~\ref{fig:distortion}.

Directly augmenting distortion in image space can create undefined areas as the image stretches to a non-rectangular shape. To resolve this, we compute a zooming factor $s$ to ensure the image can be cropped without undefined borders and incorporate this into the zooming augmentation.

\noindent \textbf{Bokeh - aperture.}
The Bokeh effect depends on the aperture $\alpha$, the depth information of the scene and a focus point $(u_\text{in}, v_\text{in})$. More precisely, larger depth distances result in a larger blur radius $b_r \in \mathbb{R}$, which can be approximately described as~\cite{peng2022bokehme,yang2016virtual} with $d_{\text{in}}$ the disparity value (invert depth) of the focus point: $b_r = \alpha \ |d - d_{\text{in}}|$.
We estimate depth~\cite{yang2024depth} and then use virtual bokeh rendering~\cite{yang2016virtual,peng2022bokehme} based on $b_r$ to augment both the aperture $\alpha$ and focus point 
$(u_\text{in}, v_\text{in})$ to simulate bokeh effects (Figure~\ref{fig:bokeh}).

\noindent \textbf{Augmentation Algorithm}
\noindent
\textit{Sampling.} In our augmentation kit, we randomly sample parameters for every frame independently. However, direct random sampling may introduce flickering artifacts as between frames the number may change too fast, which is uncommon in the real-world and can harm the learning. To prevent this, we limit the rate of change between frames and ensure smooth transitions by applying spline interpolation to the sampled parameters.

\noindent
\textit{Dropout.} We also apply augmentation dropout to ensure that: (1) a certain percentage of the original video frames are used during training, and (2) the model learns specific combinations of augmentations effectively. 

\input{tab/sota}

\noindent
\textit{Algorithm.} Our augmentation starts with bokeh using the pre-estimated depth, then applies distortion on the original images to use the maximum resolution, and followed by focal length changes, see more in the supplementary material.

%% file: fig/quals.tex
\begin{figure*}[htbp]
    \centering
    \begin{subfigure}{0.49\textwidth}
        \centering
        \includegraphics[width=\linewidth]{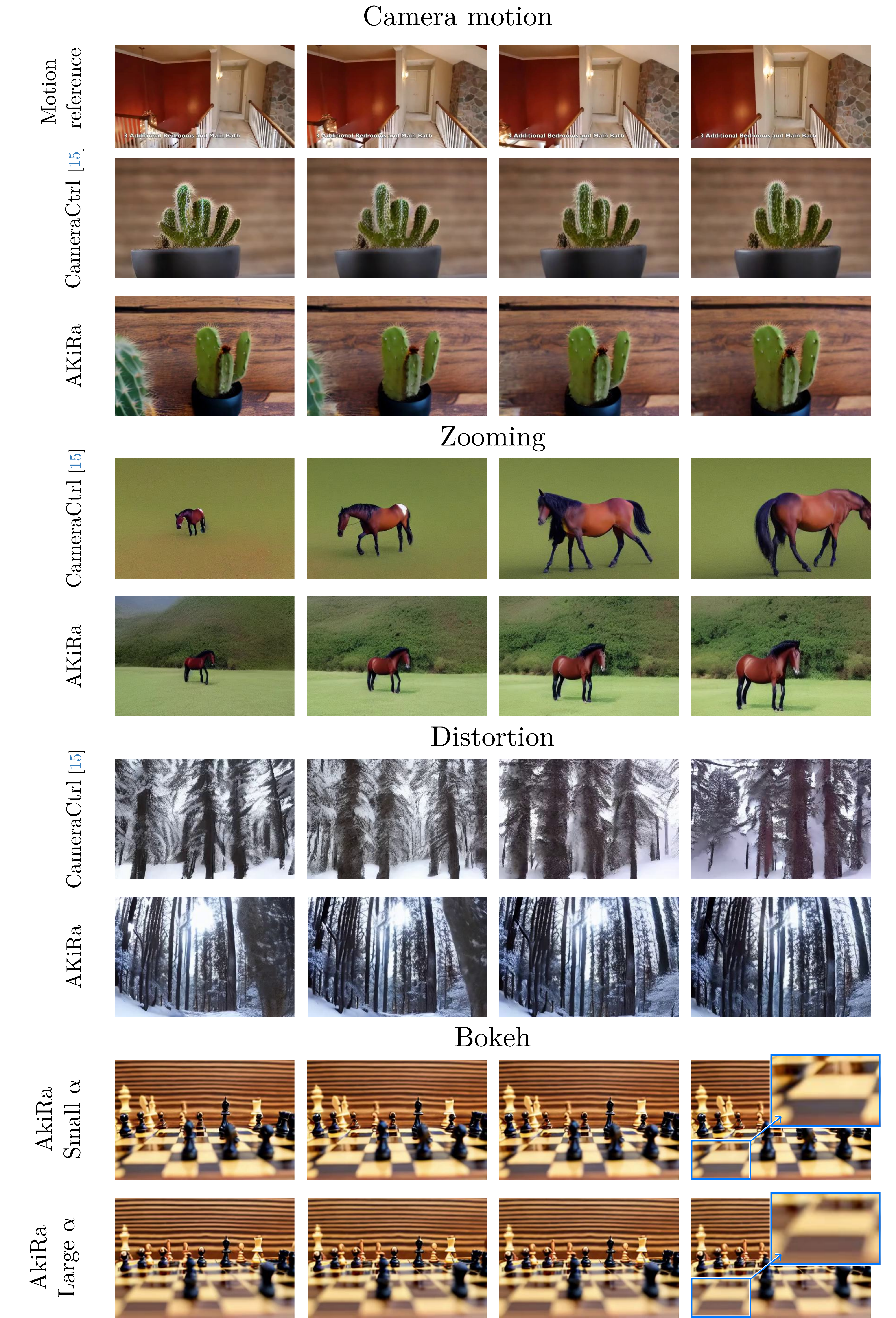}
        \caption{Animatediff~\cite{guo2023animatediff}}
        \label{fig:qual-anim}
    \end{subfigure}
    \hfill
    \begin{subfigure}{0.49\textwidth}
        \centering
        \includegraphics[width=\linewidth]{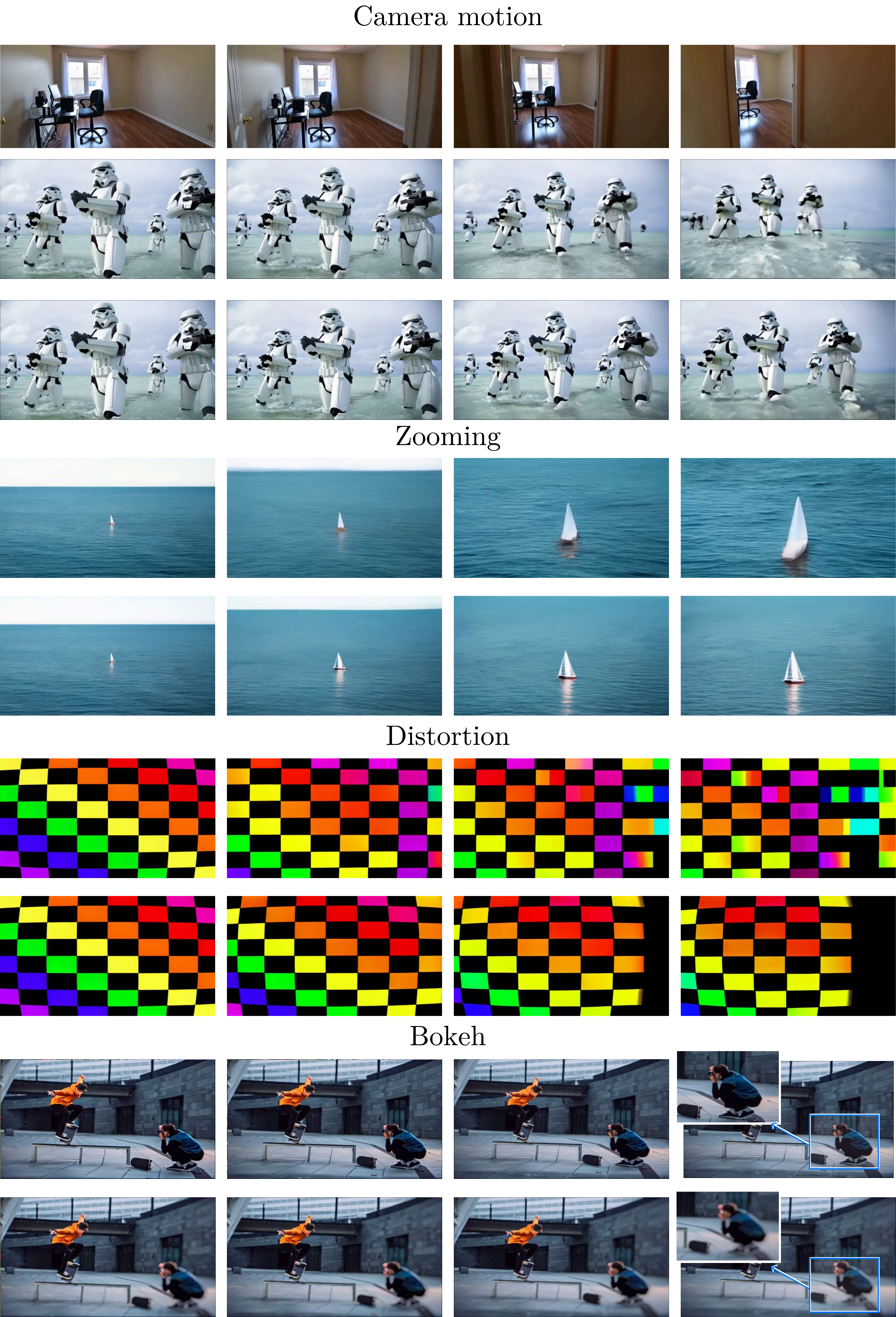}
        \caption{SVD~\cite{blattmann2023svd}}
        \label{fig:qual-svd}
    \end{subfigure}
    \caption{\textbf{Qualitative results} of \akira on Animatediff~\cite{guo2023animatediff} and SVD~\cite{blattmann2023svd} backbones. We recommend viewing the supplementary video.}
    \label{fig:quals}
\end{figure*}

%% file: tab/sota.tex
\newcolumntype{A}{>{\hspace{6pt}}c<{\hspace{6pt}}}
\newcolumntype{B}{>{\hspace{3pt}}c<{\hspace{3pt}}}
\newcolumntype{C}{>{\hspace{3pt}}c<{\hspace{3pt}}}
\newcolumntype{S}{>{\hspace{0pt}}c<{\hspace{3pt}}}

{{\renewcommand{\arraystretch}{1} 
\begin{table*}[t]
\centering
\resizebox{\textwidth}{!}{
\begin{tabular}{ll|ccc|ccc|ccc}
\toprule
\multicolumn{2}{c|}{\textbf{Method}} & \multicolumn{3}{c|}{\textbf{Video quality}} & \multicolumn{3}{c|}{\textbf{Camera motion fidelity}} & \multicolumn{3}{c}{\textbf{Dynamic consistency (VBench)}} \\ 
Backbone & Camera control & FID $\downarrow$ & FVD $\downarrow$ & CD-FVD $\downarrow$ & RPE-R (deg)$\downarrow$ &  RPE-t (cm)$\downarrow$ & FlowSim $\uparrow$ &  Consistency $\uparrow$  & Smoothness $\uparrow$ & Flickering$\uparrow$  \\ 
\midrule
\multirow{4}{*}{\rotatebox[origin=c]{60}{\shortstack{AnimateDiff\\\cite{guo2023animatediff}}}} & -  & $65.09$     & $835.2$ & $768.6$ & $4.354$  & $3.048$  & $0.00$   & $0.845$ & $0.843$ & $0.792$ \\ 
& MotionCtrl~\cite{wang2024motionctrl}                      & $104.64$ & $822.9$ & $449.7$ & \cellcolor{tabsecond}$0.443$   & $1.857$ & \cellcolor{tabsecond}$58.50$ & $0.961$ & \cellcolor{tabsecond}$0.965$ & $0.939$ \\ 
& CameraCtrl~\cite{he2024cameractrl}                        & \cellcolor{tabsecond}$61.62$  & \cellcolor{tabsecond}$384.5$ & \cellcolor{tabsecond}$355.2$ & $0.730$   & \cellcolor{tabsecond}$1.807$ & $53.74$ & \cellcolor{tabsecond}$0.965$ & $0.960$ & \cellcolor{tabsecond}$0.942$ \\ 
& AKiRa (ours)                                              & \cellcolor{tabfirst}$61.52$ & \cellcolor{tabfirst}$332.3$ & \cellcolor{tabfirst}$328.7$ & \cellcolor{tabfirst}$0.438$ & \cellcolor{tabfirst}$1.480$ & \cellcolor{tabfirst}$70.97$ & \cellcolor{tabfirst}$0.969$ & \cellcolor{tabfirst}$0.974$ & \cellcolor{tabfirst}$0.957$  \\ 
\cline{2-11}
\multirow{4}{*}{\rotatebox[origin=c]{60}{\shortstack{SVD~\cite{blattmann2023svd}}}} & - & $29.94$ & $351.0$ & $440.8$ & $1.837$ & $2.541$ & $0.56$  & $0.916$ & $0.959$ & $0.930$ \\ 
& MotionCtrl~\cite{wang2024motionctrl}           & $57.09$     & $217.3$ & $460.8$ & $1.705$    & $1.343$     & $11.96$ & $0.916$     & $0.979$     & $0.950$     \\ 
& CameraCtrl~\cite{he2024cameractrl}             & \cellcolor{tabfirst}$32.32$ & \cellcolor{tabsecond}$173.8$ & \cellcolor{tabsecond}$424.9$ & \cellcolor{tabsecond}$0.329$   & \cellcolor{tabsecond}$1.322$ & \cellcolor{tabsecond}$79.64$ & \cellcolor{tabsecond}$0.972$ & \cellcolor{tabsecond}$0.991$ & \cellcolor{tabsecond}$0.969$ \\ 
& AKiRa (ours)                            & \cellcolor{tabsecond}$32.58$     & \cellcolor{tabfirst}$162.8$      & \cellcolor{tabfirst}$398.3$      & \cellcolor{tabfirst}$0.295$    & \cellcolor{tabfirst}$1.321$     & \cellcolor{tabfirst}$80.11$      & \cellcolor{tabfirst}$0.981$     & \cellcolor{tabfirst}$0.993$     & \cellcolor{tabfirst}$0.975$     \\
\bottomrule
\end{tabular}}
\caption{\textbf{Comparison with the state-of-the-art.} 
Comparison of AKiRa and concurrent methods with different backbones on WebVid dataset, evaluating video quality, camera motion fidelity, and dynamic consistency. 
\colorbox{tabfirst}{First best} and \colorbox{tabsecond}{second best}.}
\label{tab:sota}
\end{table*}}}

%% file: sec/4.experiments.tex
\subsection{Evaluation metrics}
\label{sub:metrics}

\paragraph{For video quality} of generated videos, we report the commonly used Fréchet Inception Distance (\emph{FID})~\cite{heusel2017gans}, Fréchet Video Distance (\emph{FVD})~\cite{unterthiner2018fvd}. Given that recent works~\cite{ya2024beyond,ge2024cdfvd} point out that the \textit{FVD} tends to be biased toward content while overlooking the temporal aspects, we propose to report the Content-Debiased FVD (\emph{CD-FVD})~\cite{ge2024cdfvd}, considered more relevant.

\noindent \textbf{Evaluating camera motion fidelity} is challenging for generated content. Some approaches use SLAM and pose estimation methods~\cite{schonberger2016structure,pan2024glomap,zhao2022particlesfm} to evaluate the estimated trajectory from generated video content (MotionCtrl~\cite{wang2024motionctrl}, CameraCtrl~\cite{he2024cameractrl}). However, these often assume (partial) static consistency, which is hard to maintain in the generated content due to flickering or unrealistic motion artefacts but irrelevant to motion quality; (ii) the absence of precise camera model definitions (e.g., focal length, distortion) further complicates trajectory-based assessments~\cite{sturm2012benchmark,jinyu2019survey} given that optic parameters greatly influence motion interpretation; 
(iii) trajectory-based methods are computationally intensive (can reach {$\sim$}10 minutes per $16$-frame video~\cite{zhao2022particlesfm}) and struggle to scale efficiently for large-scale video assessments. See motion evaluation in supplementary materials.

Therefore, to show the \textbf{camera motion fidelity towards the control} we report two metrics: first, following~\cite{wang2024motionctrl, he2024cameractrl} reporting trajectory errors, we estimate dense camera poses using ParticleSfM~\cite{zhao2022particlesfm} but calculate scale-corrected rotational and translational relative pose errors between frames (\emph{RPE-R} and \emph{RPE-t}, respectively). The choice of using scale correction and relative pose aims at reducing the unstable estimation of trajectory-based metrics~\cite{sturm2012benchmark, jinyu2019survey}.

Additionally, to measure \textbf{motion alignment} between reference and generated videos we propose a flow similarity metric (\emph{FlowSim}). This metric relies less on camera model parameters and focuses primarily on frame-to-frame motion. Moreover, the optical flow-based approach is computationally efficient and scalable with GPU parallelism. For this, we estimate the optical flow of both videos $\mathbf{F}_{r}$ and $\mathbf{F}_{g}$ using RAFT~\cite{teed2020raft}, then extract the flow magnitude $\Vert \mathbf{F} \Vert$ and direction $\boldsymbol{\varphi}$. To filter out residual noise, we consider only flow components with a magnitude above a set threshold $t$, and we compute the cosine similarity of the directional components as follows, with $\mathds{1}$ an indicator function:
\begin{equation}
    \text{FlowSim}(\mathbf{F}_{r}, \mathbf{F}_{g}) = \mathds{1}_{\Vert\mathbf{F}_{r}\Vert \& \Vert\mathbf{F_{r}}\Vert > t} \cdot  \boldsymbol{\varphi}_{r} \cdot  \boldsymbol{\varphi}_{g} \ ,
    \label{eq:flowsim}
\end{equation}

\noindent \textbf{Dynamic consistency (VBench).} We also assess the \textbf{temporal dynamic consistency} of generated videos with the subject consistency (\emph{Consistency}), smoothness (\emph{Smoothness}) and temporal flickering (\emph{Flickering}) metrics proposed in the video benchmark suite VBench~\cite{huang2024vbench}.

\noindent \textbf{Optical consistency.} To evaluate zoom (\emph{ZoomSim}) and distortion (\emph{DistortSim}), we use \emph{FlowSim} to measure the similarity between generated and theoretical optical flow (see supplementary materials).\ For evaluating bokeh effects, inspired by~\cite{courant2021high},  we use an off-the-shelf defocus detector~\cite{zhao2023defocus} and report the in-focus area (\emph{FocusArea}) across varying aperture levels: a low aperture results in a wide in-focus area (everything appears in focus), while a high aperture produces a narrow in-focus area, showing selective focus.

\subsection{Quantitative comparison}
\label{sec:quantitative_comp}
\input{tab/vic_user_ablation_extra_figure}

\noindent \textbf{Comparison to the state of the art.} Table~\ref{tab:sota} reports the performance of  AKiRa against state-of-the-art camera control approaches for video generation: MotionCtrl~\cite{wang2024motionctrl} and CameraCtrl~\cite{he2024cameractrl}. 
We evaluate video quality, camera motion fidelity and temporal dynamic consistency metrics (Section~\ref{sub:metrics}) for text-to-video (T2V) backbone Animatediff~\cite{guo2023animatediff} and image-to-video (I2V) backbone SVD~\cite{blattmann2023svd}, also reporting baselines without any camera control module (\textit{-}).
All metrics are computed on $1,000$ generated samples using random text prompts or conditioning frames from the WebVid dataset~\cite{bain2021webvid}. For \akira, as we explicitly control the bokeh, to improve the realism of the generated results, we set $\alpha=50$ (see discussion in supplementary materials).

\noindent \textbf{In video quality,} AKiRa achieves leading performance across quality, camera motion fidelity, and dynamic consistency metrics, outperforming other state-of-the-art methods. Especially, for FVD and CD-FVD, AKiRa scores best with both AnimateDiff ($332.3$, $328.7$) and SVD ($162.8$, $398.3$), outperforming CameraCtrl's $384.5$ and $355.2$ on AnimateDiff, and $173.8$ and $424.9$ on SVD. 

\noindent \textbf{For camera motion fidelity,} we observe that: (1) AKiRa achieves the highest FlowSim scores (AnimateDiff: $70.97$; SVD: $80.11$), ahead of MotionCtrl $58.50$ with AnimateDiff, and CameraCtrl $79.64$ with SVD, and lowest motion errors, both translational and rotational ($0.438$ and $1.480$ on Animatediff); (2) the proposed metric, FlowSim, assigned close to zero values to random motion, thus proportionally reflecting camera pose errors with improved interpretability. 

\noindent \textbf{For dynamic consistency,} AKiRa leads in VBench scores, e.g. achieving $0.969$, $0.974$, $0.957$ on AnimateDiff, and $0.981$, $0.993$, $0.975$ on SVD. These results validate that disentangling camera components during training enables high-quality and consistent video synthesis.

\noindent \textbf{Optical comparison.} In Table~\ref{tab:optical}, we show the optical consistency performances (Section~\ref{sub:metrics}). We focus on CameraCtrl~\cite{he2024cameractrl} since its ray-based representation supports zoom and distortion effects, though does not account for bokeh, whereas MotionCtrl~\cite{wang2024motionctrl} lacks these effects due to its camera model. We also report raw Animatediff~\cite{guo2023animatediff} as the baseline without any control module (\textit{-}). 
We generate dedicated datasets of 1,000 samples for each category. For zoom and distortion, we create static motion videos with random zoom-in/zoom-out effects and distortions. For bokeh, we generate 1,000 samples with varying apertures ($\alpha$=$0$/$30$/$100$).
Table~\ref{tab:optical} highlights that \akira outperforms both AnimateDiff and CameraCtrl in optical coherence metrics, achieves the highest \emph{ZoomSim} and \emph{DistortSim} scores ($86.82$ and $81.19$), showing superior zoom and distortion handling compared to CameraCtrl. In terms of bokeh, we observe a decreasing trend of the \emph{FocusArea} with increasing aperture $\alpha$ from $0$ to $100$, confirming the controllability of the aperture feature (a larger aperture results in a bigger bokeh area). 
These results highlight AKiRa’s ability to control zoom, distortion, and focus effects.

\noindent \textbf{User study.} We conducted a user study with $25$ participants, evaluating (1) video, (2) text-to-video, (3) motion, (4) zoom, (5) distortion effect, and (6) bokeh effect qualities. Participants viewed $10$ samples, with each generated by: the baseline (no camera control module), MotionCtrl~\cite{wang2024motionctrl}, CameraCtrl~\cite{he2024cameractrl}, and our AKiRa, both the Animatediff~\cite{guo2023animatediff} and SVD~\cite{blattmann2023svd} backbones. For each criterion, participants ranked the methods from $1$ (best) to $4$ (worst). 

In Table~\ref{tab:user-study} we report the proportion of participants ranking each method as 1st across various criteria (with $25\%$ as a random baseline). Results show that \akira consistently ranks highest, with an average ranking proportion of $44.2\%$, compared to $29.2\%$ for the next best method, CameraCtrl~\cite{he2024cameractrl}. Specifically, $43.6\%$ of participants preferred CameraCtrl's zoom effect, followed closely by \akira at $35.6\%$. This preference likely stems from CameraCtrl's overly fast and impactful zoom effect, though it lacks subject consistency upon closer inspection (see Section~\ref{sub:quals} and Figure~\ref{fig:quals}). Overall, \akira notably enhances the aesthetic quality of the backbone, with $39.6\%$ and $55.2\%$ of participants ranking it 1st for video quality and text-video consistency, compared to only $13.2\%$ and $11.2\%$ for the baselines.

\noindent \textbf{Ablation study.} In Table~\ref{tab:ablation}, we analyze the impact of each augmentation in \akira on overall model performance. For each augmentation combination, we train a model using the Animatediff~\cite{guo2023animatediff} backbone. The results show that applying all augmentations yields the best performances in both visual (\emph{CD-FVD}) and motion (\emph{FlowSim}) quality. The strong performance of distortion with bokeh (second-to-last row) is due to the cropping and rescaling process intrinsic to distortion augmentation (see Sec.~\ref{sec:method}), which implicitly works as a focal length adjustment, leading to comparable results. Additionally, the high \emph{FlowSim} score for bokeh-only augmentation (fourth row) might be due to overly smoothed content, which implicitly improves the similarity measure.

\subsection{Qualitative results}
\label{sub:quals}

\noindent \textbf{Qualitative comparison.} In Figure~\ref{fig:quals} we present a qualitative comparison of the various effects—motion, zoom, distortion, and bokeh—between \akira and CameraCtrl~\cite{he2024cameractrl}, using both the Animatediff~\cite{guo2023animatediff} and SVD~\cite{blattmann2023svd} backbones. 
For motion (first part), \akira accurately reproduces reference motion from the RealEstate dataset~\cite{zhou2018realestate} across both backbones. 
In zooming (second part), our method demonstrates better consistency, while CameraCtrl struggles; for example, in the SVD examples (second part, second column), \akira maintains the boat's position and orientation, unlike CameraCtrl, where the boat shows deformation.
In terms of distortion (third part), CameraCtrl fails to achieve this effect, whereas \akira successfully applies it; for instance, in the Animatediff example (third part, first column), the tree trunks visibly distort. Lastly, for the bokeh effect (fourth part), \akira adds adjustable out-of-focus blur. In the Animatediff example (fourth part, first column), for instance, the foreground of the chessboard displays varying blur levels depending on the aperture.

\noindent \textbf{A special case: dolly zoom.} 
A dolly zoom is an iconic cinematic effect that produces a dramatic perspective shift, achieved by simultaneously pulling while zooming in reverse direction. 
We show in Figure~\ref{fig:dolly} that \akira successfully reproduces the dolly zoom effect, whereas CameraCtrl~\cite{he2024cameractrl} achieves only a pure zoom-out effect. 
This is due to \akira’s precise disentangling of key camera parameters, especially focal length and camera position. We recommend viewing the supplementary video.

%% file: tab/vic_user_ablation_extra_figure.tex

\begin{figure*}[htbp]
\centering
  \begin{minipage}[b]{0.55\textwidth}
    
    \centering
    \includegraphics[width=\linewidth]{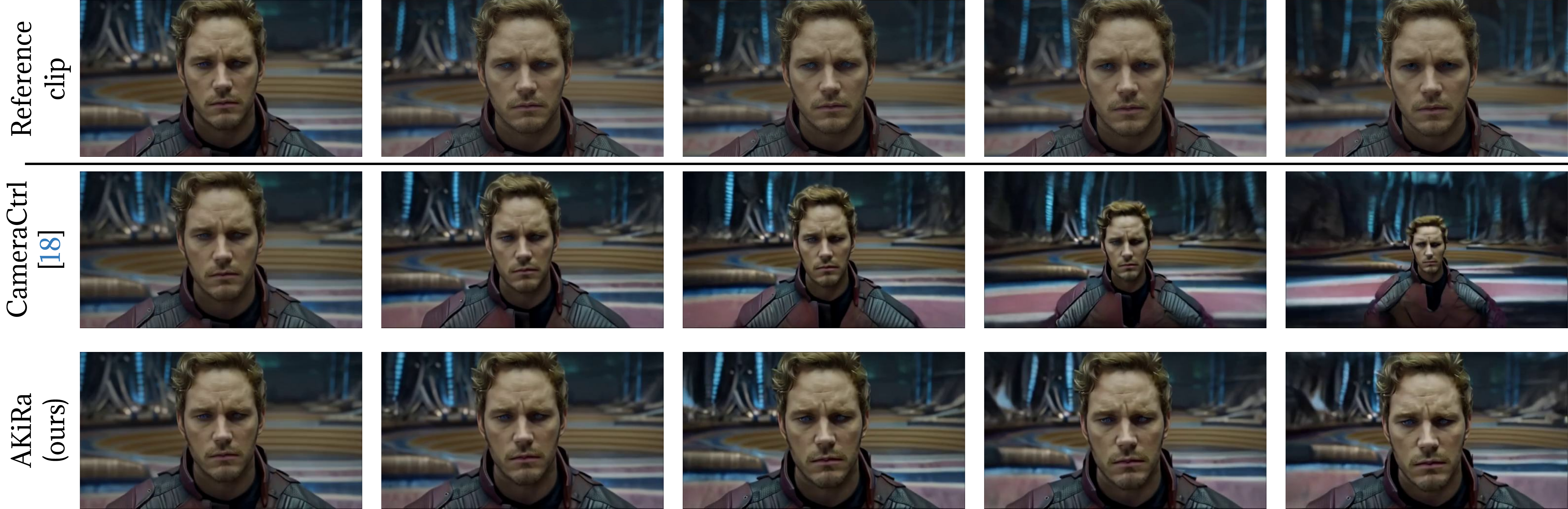}
    \captionof{figure}{\textbf{Comparison of dolly zoom} effect on SVD (I2V)~\cite{blattmann2023svd} backbone between \akira, CameraCtrl\cite{he2024cameractrl} and the reference from \emph{Guardians of the Galaxy Vol. 2} (\emph{Gunn, 2017}). Note that the background recedes while the character remains static.
    \label{fig:dolly}}
    \centering
    \footnotesize{
    \resizebox{\textwidth}{!}{
    \input{tab/user_study}  
    }}
    \captionof{table}{\textbf{Results of user study.} \% participants ranking each method as 1st for:  \emph{Video} quality, \emph{Text-to-video} consistency, motion fidelity, and optical coherence (\emph{Zoom}, \emph{Distortion}, \emph{Bokeh}). \colorbox{tabfirst}{First best} and \colorbox{tabsecond}{second best}.
    }
    \label{tab:user-study} 
    
%
  \end{minipage}
  \hfill
  \begin{minipage}[b]{0.43\textwidth}
    \footnotesize{
        \resizebox{\textwidth}{!}{
        {\begin{tabular}{@{}l ccc@{\hspace{3pt}}c@{\hspace{3pt}}c@{}}
        \toprule
        \multirow{2}{*}{\textbf{Method}}                & \multirow{2}{*}{\textbf{ZoomSim} $\uparrow$} & \multirow{2}{*}{\textbf{DistortSim} $\uparrow$} & \multicolumn{3}{c}{\textbf{FocusArea}} \\ 
              &  &   & $\alpha$=$0$ & $\alpha$=$30$ & $\alpha$=$100$ \\
        \midrule
        AnimateDiff~\cite{guo2023animatediff} &                     $0.00$ &                      $0.00$ & $0.90$ & - & -\\
        CameraCtrl~\cite{he2024cameractrl}   &                     $78.98$ &                     $62.13$ & $0.69$ & - & -\\
        AKiRa (ours)                         & $86.82$ & $81.19$ & $0.72$ & $0.64$ & $0.61$ \\
        \bottomrule 
        \end{tabular}}}}
        \captionof{table}{\textbf{Optical comparison} of AKiRa and CameraCtrl~\cite{he2024cameractrl} with Animatediff~\cite{guo2023animatediff} backbone on WebVid, evaluating optical coherence (zoom, distortion and bokeh). }
        \label{tab:optical}
    \centering
    \input{tab/ablation}

    \captionof{table}{\textbf{Ablation of augmentations.} Ablation of each augmentation combination in \akira. All models are trained on Animatediff~\cite{guo2023animatediff} backbone.}
    \label{tab:ablation}
  \end{minipage}
  
\end{figure*}

%% file: tab/user_study.tex
\begin{tabular}{@{}lccccccc@{}}
\toprule
\textbf{Method} & \textbf{Video} & \textbf{Text} & \textbf{Motion} & \textbf{Zoom} & \textbf{Distortion} & \textbf{Bokeh} & \textbf{Average} \\ \midrule
Baseline~\cite{guo2023animatediff,blattmann2023svd} & $13.2$ & $11.2$ &  $7.2$ &  $4.0$ &  $8.4$ &  $7.2$ &  $8.5$ \\
MotionCtrl~\cite{wang2024motionctrl}                & \cellcolor{tabsecond}$27.6$ &  $6.4$ & $21.6$ & $16.8$ & $18.8$ & $17.6$ & $18.1$ \\
CameraCtrl~\cite{he2024cameractrl}                  & $19.6$ & \cellcolor{tabsecond}$27.2$ & \cellcolor{tabsecond}$23.6$ & \cellcolor{tabfirst}$43.6$ & \cellcolor{tabsecond}$29.6$ & \cellcolor{tabsecond}$31.6$ & \cellcolor{tabsecond}$29.2$ \\
AKiRa (ours)                                        & \cellcolor{tabfirst}$39.6$ & \cellcolor{tabfirst}$55.2$ & \cellcolor{tabfirst}$47.6$ & \cellcolor{tabsecond}$35.6$ & \cellcolor{tabfirst}$43.2$ & \cellcolor{tabfirst}$43.6$ & \cellcolor{tabfirst}$44.2$ \\ \bottomrule
\end{tabular}

%% file: tab/ablation.tex
\begin{tabular}{@{}c@{\hspace{4pt}}c@{\hspace{3pt}}c@{\hspace{3pt}}|c@{\hspace{5pt}}c@{}}
\toprule
\textbf{Focal} & \textbf{Lens} & \textbf{Aperture}                    & \textbf{CD-FVD} $\downarrow$    &  \textbf{FlowSim} $\uparrow$   \\
\midrule
           &            &            &  $355.2$ & $53.74$ \\
\checkmark &            &            &  $351.2$ & $67.25$ \\
           & \checkmark &            & $360.1$ & $71.82$ \\
           &            & \checkmark &  $375.8$   & $73.27$ \\
\checkmark & \checkmark &            &  $359.4$ & $66.86$ \\
\checkmark &            & \checkmark &  $332.7$ & $62.93$ \\
           & \checkmark & \checkmark &  $331.0$ & $71.97$ \\
\checkmark & \checkmark & \checkmark & $328.7$ & $70.97$ \\ 
\bottomrule
\end{tabular}

%% file: sec/5.conclusion.tex
In this paper, we introduce the concept of optical video generation, a framework that allows users to control camera motions as well as optical parameters. 
We trained a dedicated camera adapter using a set of data augmentation techniques —\akira— that pairs camera/lens parameters with corresponding videos. 
Results show that our framework generates optically coherent content, outperforming state-of-the-art approaches while offering extra control. 
\akira expands the possibilities of video generation and bridges the gap between synthetic and real-world capabilities.

%% file: sec/acknowledgements.tex
This work was supported by ANR-22-CE23-0007, ANR-22-CE39-0016, Hi!Paris grant and fellowship, and was granted access to the High-Performance Computing (HPC) resources of IDRIS under the allocation 2024-AD011014300R1 made by GENCI. 
We would like to thank Hongda Jiang and Yuanzhi Zhu for their insightful comments and suggestions.

%% file: supmat.tex




%
\definecolor{cvprblue}{rgb}{0.21,0.49,0.74}




\maketitlesupplementary

\noindent In this supplementary material, we include:

\begin{enumerate}[label=\Alph*.~]
    
    \item \nameref{supsec:quants} 
    \item \nameref{supsec: aperture} 
    \item \nameref{supsec:ethical} 
    \item \nameref{supsec: algo}
    \item \nameref{supsec:user-study}
    \item \nameref{supsec:metrics}
    \item \nameref{supsec:quali} 
\end{enumerate}

\renewcommand{\thesection}{\Alph{section}}

\section{Additional quantitative results}
\label{supsec:quants}
In this section, we present additional quantitative results on the RealEstate10K dataset~\cite{zhou2018realestate}, comparing the performance of AKiRa with state-of-the-art camera control approaches for video generation: MotionCtrl~\cite{wang2024motionctrl} and CameraCtrl~\cite{he2024cameractrl}, incorporating the corresponding LoRA~\cite{hu2022lora} module to ensure domain consistency. 

We evaluate video quality, camera motion fidelity, and temporal dynamic consistency metrics (same as in Table~\ref{tab:sota} of the main manuscript) for the text-to-video (T2V) backbone Animatediff~\cite{guo2023animatediff} and the image-to-video (I2V) backbone SVD~\cite{blattmann2023svd}. 

All metrics are computed on 1,000 generated samples using text prompts from the RealEstate10K dataset or conditioning frames from the WebVid dataset~\cite{bain2021webvid}. For AKiRa, as it explicitly controls the bokeh, we set \(\alpha = 0\) to align with the aperture behavior of the RealEstate10K dataset, where videos are recorded using small apertures and wide-angle cameras. The impact of aperture on performance is discussed in Section~\ref{supsec: aperture} and detailed in Table~\ref{supptab:aperture}.

In Table~\ref{supptab:aperture}, we present the performance of AKiRa in comparison with two state-of-the-art methods: MotionCtrl~\cite{wang2024motionctrl} and CameraCtrl~\cite{he2024cameractrl}. In terms of video quality, AKiRa outperforms both SoTA methods on FVD and CD-FVD metrics for both the Animatediff and SVD backbones, achieving scores of 128.55 and 89.16 for Animatediff, and 54.83 and 41.55 for SVD. A similar trend is observed in dynamic consistency, where AKiRa leads across all metrics on both backbones. It achieves the highest scores for Consistency (0.9851), Smoothness (0.9933), and Flickering (0.9733), demonstrating superior temporal coherence. In terms of motion fidelity, AKiRa demonstrates significantly better motion controllability on the Animatediff backbone, achieving superior performance as evidenced by the highest FlowSim and lowest RPE errors. AKiRa competes closely with CameraCtrl on SVD, with only a narrow difference. We attribute this to the overfitting of CameraCtrl on the real-estate dataset, where intrinsic parameters and optical features remain unchanged during this experiment.

\input{tab/supp-sota}

\section{Additional analysis of bokeh map }
\label{supsec: aperture}

\input{fig/supmat_bokeh}

\paragraph{Controllability of Bokeh - Aperture}
We propose a bokeh map with the same structure and dimensions as the direction and moment maps, assigning an aperture and focus (depth-of-field) parameter 
to each pixel in the frame. 
More specifically, we define the coordinates of the focus point $(u_\text{in}, v_\text{in})$ representing the sharpest point in the frame. We then define the per-pixel bokeh map as $\mathbf{a} \in \mathbb{R}^3$ for any point $(u, v)$ on the frame as:

\begin{equation}
    \mathbf{a} = \begin{bmatrix} u - u_\text{in} \\ v - v_\text{in} \\ \Vert (u, v) - (u_\text{in}, v_\text{in})\Vert^{\frac{1}{\sigma(\alpha)}} \end{bmatrix}\ ,
\label{supeq: aperture}
\end{equation}

We first present the visualization of the bokeh map and demonstrate how it influences the generated videos. We examine two groups of bokeh variations: the effect of varying the aperture \(\alpha\) in Figure~\ref{fig:qual-bokeh-aperture}, and the effect of adjusting the focus point \(f_{in}\) in Figure~\ref{fig:qual-bokeh-focus-point} (see the zoomed image in the second row highlighted the red rectangle). 

In Figure~\ref{fig:qual-bokeh-aperture}, we progressively increase the aperture level over time, with the focus fixed at the center of the image. As a result, the visualization of the bokeh map shrinks, and it shows that the blur area expands proportionally with increasing aperture \(\alpha\).

In Figure~\ref{fig:qual-bokeh-focus-point}, we shift the focus point \(f_{in}\) from the upper-left corner to the lower-right corner. This causes the center of the bokeh map to move accordingly, effectively shifting the blur area in generated videos. The results confirm our ability to dynamically control the blur area based on the focus point.

Both experiments validate the effectiveness and controllability of \akira in manipulating the depth of field.

\paragraph{Bokeh influence on T2V performance}
We analyze the influence of bokeh on video quality, flow similarity, and dynamic consistency using Animatediff, with the experimental settings identical to those in Table~\ref{tab:sota} of the main manuscript. By varying the aperture value \(\alpha\) from 0 to 100, we measure its impact on the corresponding metrics.

Table~\ref{supptab:aperture} reports the results for different aperture settings. When the aperture is small and bokeh is weak (i.e., \(\alpha=0\)), the generated videos exhibit better consistency. Conversely, when the aperture is large (i.e., \(\alpha=100\)), the videos demonstrate greater smoothness and reduced flickering. Notably, the optimal video generation metrics are observed around \(\alpha = 30\) and \(\alpha = 50\). This can be attributed to \textbf{the intrinsic bokeh} present in the WebVid~\cite{bain2021webvid} dataset, where \textbf{adding an appropriate amount of bokeh enhances realism}, resulting in improved FVD and CD-FVD performance.

\input{tab/supp-aperture}

\section{Ethical discussion}
\label{supsec:ethical}
Our paper proposes a method to generate videos based on camera and optical control, enabling better alignment of the generation process with user intentions and geometric information. On the positive side, this approach can enhance the AIGC (AI-Generated Content) creative process by reducing biases introduced by training data, more accurately reflecting user intentions, and minimizing trial-and-error in content generation. This efficiency can also contribute to reducing the carbon footprint associated with the generation process. On the negative side, however, it may reduce the labour required for video production, potentially leading to job losses, and could stifle creativity if individuals become overly reliant on generative tools.

\section{Algorithm of AKiRa}
\label{supsec: algo}
We present the complete AKiRa algorithm in Algorithm~\ref{alg:akira}. As discussed in the main manuscript, random sampling is implemented using spline sampling. Augmentation dropout with a probability \(p\) is applied to all the optical features, both collectively and individually.

Since augmentation is performed on-the-fly during the training process, the augmentation order is carefully designed to optimize computational efficiency. Specifically, we first augment the bokeh aperture to leverage the pre-computed depth map derived from the original frames. Next, distortion augmentation is applied, which may implicitly alter the focal length due to a necessary cropping operation to avoid undefined borders during image warping. Finally, we augment the zoom aspect, incorporating the results of the distortion augmentation.

During training, the augmentation parameters are sampled as follows: the dropout probability \(p\) is set to \(0.2\); the bokeh aperture is sampled uniformly between \(0\) and \(100\); the distortion parameters are sampled uniformly within the range \([-0.1, 0.1]\) for all three parameters in $\mathbf{D}$; and the zoom factor is sampled between \(1.0\) and \(3.0\), as zoom factors below \(1.0\) are ill-defined due to the difficulty to generate outpainting content through augmentation.

\begin{algorithm}
\caption{\akira augmentation algorithm}\label{alg:akira}
\begin{algorithmic}
\Require $\mathbf{I}$: frames, $\mathbf{Z}$: depth maps, $p$: aug. dropout.
\If{True with probability $p$} 
    \If{True with probability $p$}
        \State $\{\alpha\} = \textsc{RandomApertureSpline}()$ 
        \State $\{(u, v)\} = \textsc{RandomInfocusSpline}()$ 
        \State $\mathbf{I} \gets$ \textsc{BokehAugmenter}(${\mathbf{I}}$, ${\mathbf{Z}}$, ${\{\alpha\}}$, {$\{(u, v)\}$})
    \EndIf
    \If{True with probability $p$}
        \State $\{\mathbf{D}\} = \textsc{RandomDistortionSpline}()$ 
        \State $\mathbf{I} \gets$ \textsc{DistortionAugmenter}($\mathbf{I}, \{\mathbf{D}\}$)
    \EndIf
    \If{True with probability $p$}
        \State $\{f\} = \textsc{RandomFocalSpline}()$ 
        \State $\mathbf{I} \gets$ \textsc{ZoomAugmenter}($\mathbf{I}$, \{f\})
    \EndIf

    \Return  $\mathbf{I}$
\Else{}
    \Return  $\mathbf{I}$
\EndIf
\end{algorithmic}
\end{algorithm}

\section{Details about user study}
\label{supsec:user-study}

In this section, we elaborate on the specifics of our user study setup corresponding to Section~\ref{sec:quantitative_comp} in our main manuscript.

For the evaluation, each participant was presented with a total of $10$ video sets, $5$ with Animatediff backbone (T2V) and $5$ with SVD backbone (I2V). 
Each set comprised $4$ generated videos from (i) the baseline (backbone without camera control), (ii) MotionCtrl, (iii) CameraCtrl, and (iv) AKiRa (ours), we shuffled the results and displayed them in random order.

Subsequently, participants were prompted with $6$ questions for each comparison:
\begin{enumerate}
    \item \textit{Rank the consistency of the video with the text prompt} (Only for Animatediff backbone).
    \item \textit{Rank the video quality (i.e. temporal consistency).}
    \item \textit{Rank the camera motion consistency with the reference.}
    \item \textit{Rank the best zoom-in or -out effect.}
    \item \textit{Rank the best distortion effect (e.g. fisheye).}
    \item \textit{Rank the best bokeh (in- or out-of-focus effect).}
\end{enumerate}

In total, we recorded $25$ participants with each participant responding to $55$ questions. We analyzed the results by examining responses to each question individually, summarizing the collective feedback. 

\section{Discussion about metrics}
\label{supsec:metrics}

\subsection{Drawbacks of SfM-based metrics}
\begin{figure}
    \centering
    \includegraphics[width=\linewidth]{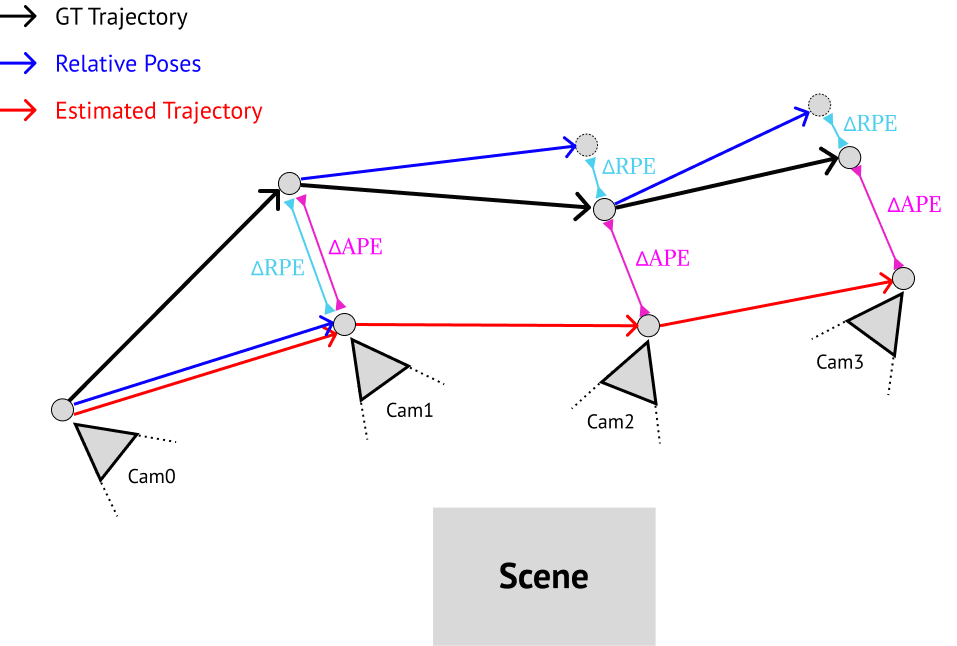}
    \caption{\textbf{Illustrates the instability of Absolute Pose Error (APE), purple} when measuring trajectory accuracy, compared to the robustness of Relative Pose Error (RPE), cyan. APE tends to accumulate errors, especially at later frames. In contrast, RPE calculates errors based on relative transformations between consecutive frames, making it less sensitive to single-frame errors and more robust for trajectory evaluation.}
    \label{fig: pose metric}
\end{figure}

\paragraph{Absolute Pose Error}
Many recent works~\cite{xu2024camco,wang2024motionctrl,he2024cameractrl} on camera motion-controlled video generation rely on Structure-from-Motion (SfM) or SLAM-based metrics to evaluate the effectiveness of camera control. The primary metric utilized in these works is similar to the Absolute Pose Error (APE) used in SLAM and SfM applications~\cite{schonberger2016structure,sturm2012benchmark}, which computes the average trajectory error for translation and rotation separately. These errors are defined as follows:

For translation error in \(\mathbb{R}^3\):
\[
\text{APE}_{\text{trans}} = \frac{1}{N} \sum_{i=1}^N \| \hat{\mathbf{t}}_i - \mathbf{t}_i^* \|,
\]
where \(N\) is the total number of frames, \(\hat{\mathbf{t}}_i \in \mathbb{R}^3\) is the estimated camera translation vector for frame \(i\), and \(\mathbf{t}_i^* \in \mathbb{R}^3\) is the ground truth translation vector for the same frame.

For rotation error in \(SO(3)\), the error is often computed as the angle of the relative rotation:
\[
\text{APE}_{\text{rot}} = \frac{1}{N} \sum_{i=1}^N \arccos \left( \frac{\text{trace}\left( \hat{\mathbf{R}}_i (\mathbf{R}_i^*)^\top \right) - 1}{2} \right),
\]
where \(\hat{\mathbf{R}}_i \in SO(3)\) is the estimated rotation matrix for frame \(i\), and \(\mathbf{R}_i^* \in SO(3)\) is the ground truth rotation matrix for the same frame.
However, APE is highly sensitive to errors in individual frames, which is a common issue in generated videos due to \textbf{flickers or sudden object movements}. These artifacts, often \textbf{irrelevant to the quality of motion control}, can disproportionately affect the evaluation. 

\begin{figure*}[h]
    \centering
    \begin{subfigure}{0.49\textwidth}
        \centering
        \includegraphics[width=\linewidth]{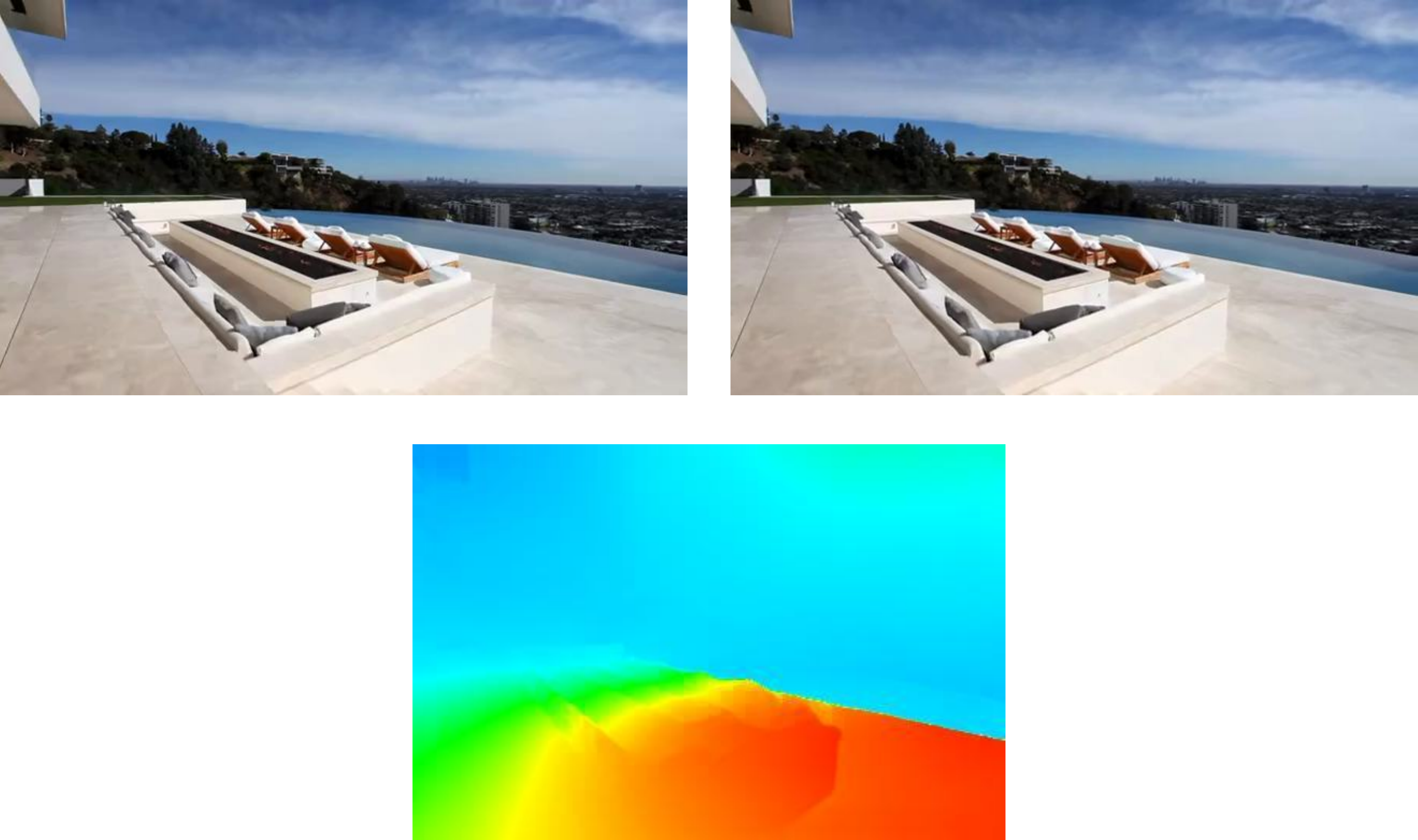}
        \caption{Ground-truth optical flow.}
    \end{subfigure}
    \hfill
    \begin{subfigure}{0.49\textwidth}
        \centering
        \includegraphics[width=\linewidth]{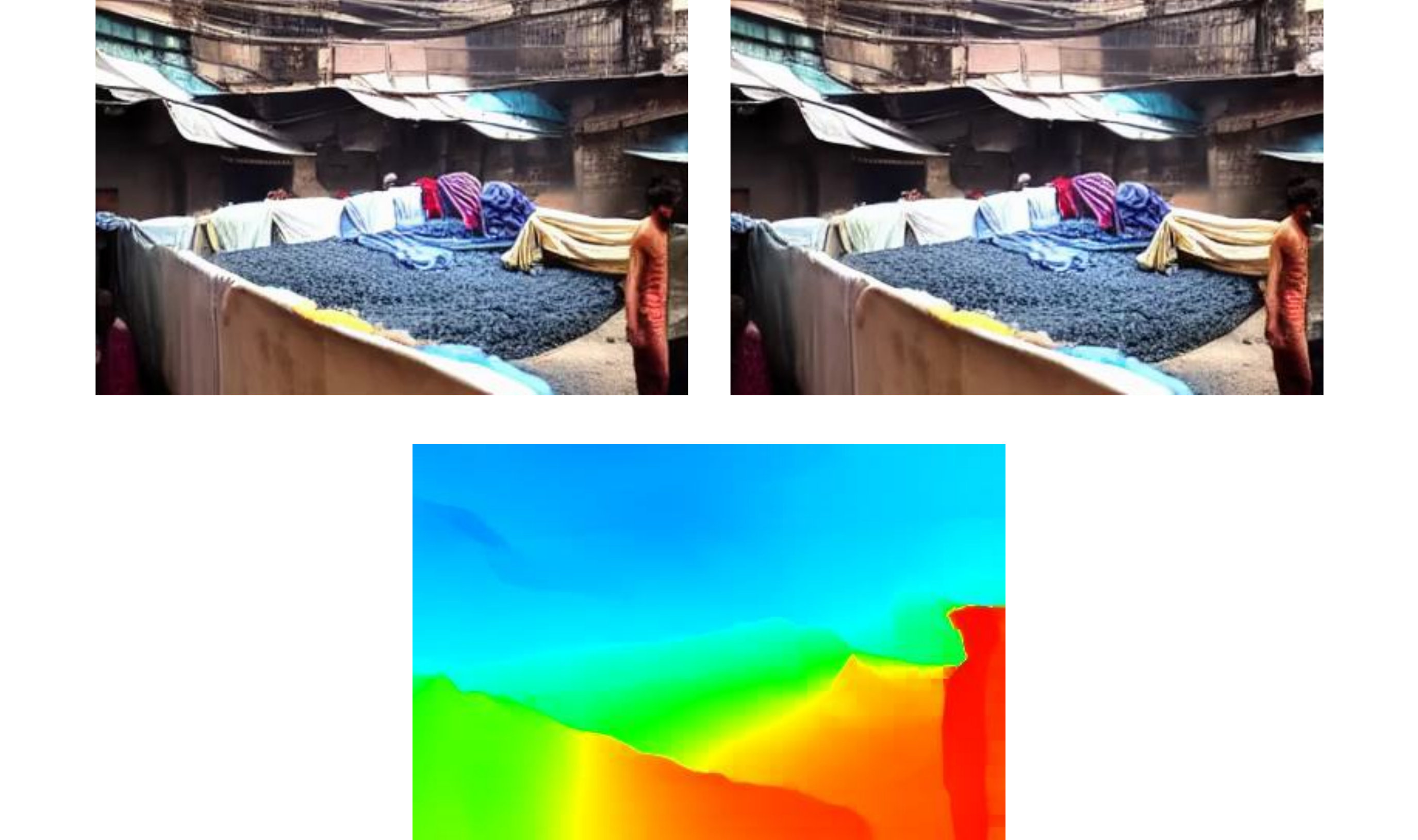}
        \caption{Generated optical flow}
    \end{subfigure}
    \caption{\textbf{Comparison of optical flow} for different content and similar camera motion.} 
    \label{fig:supp-flow}
\end{figure*}

\paragraph{Relative Pose Error}
To address the APE limitation, related domains often rely on the Relative Pose Error (RPE), which reduces the impact of \textbf{accumulated errors} caused by single-frame inaccuracies, especially when such errors occur at the early stages of the trajectory.

RPE is computed by comparing the relative transformations between consecutive frames, rather than the absolute poses. It is defined separately for translation and rotation as follows:

For translation error in \(\mathbb{R}^3\):
\[
\text{RPE}_{\text{trans}} = \frac{1}{N-1} \sum_{i=1}^{N-1} \| \Delta\hat{\mathbf{t}}_i - \Delta\mathbf{t}_i^* \|,
\]
where \(\Delta\hat{\mathbf{t}}_i = \hat{\mathbf{t}}_{i+1} - \hat{\mathbf{t}}_i\) is the estimated relative translation, and \(\Delta\mathbf{t}_i^* = \mathbf{t}_{i+1}^* - \mathbf{t}_i^*\) is the ground truth relative translation.

For rotation error in \(SO(3)\), the relative error is defined as:
\[
\text{RPE}_{\text{rot}} = \frac{1}{N-1} \sum_{i=1}^{N-1} \arccos \left( \frac{\text{trace}\left( \Delta\hat{\mathbf{R}}_i \Delta\mathbf{R}_i^* \right) - 1}{2} \right),
\]
where \(\Delta\hat{\mathbf{R}}_i = \hat{\mathbf{R}}_{i+1} \hat{\mathbf{R}}_i^\top\) is the estimated relative rotation, and \(\Delta\mathbf{R}_i^* = \mathbf{R}_{i+1}^* \mathbf{R}_i^{*^\top}\) is the ground truth relative rotation.

We demonstrate this phenomenon in Figure~\ref{fig: pose metric}. When the first frame exhibits a high APE (purple), even if the subsequent trajectory is relatively accurate, the error is accumulated throughout the trajectory. In contrast, RPE is computed between relative poses (cyan), \textbf{making it less biased by errors in previous estimations}, therefore providing a more robust assessment of motion control quality.

\input{fig/supp-quals}

\paragraph{Scaling Ambiguity}
Another challenge when using 3D metric errors to evaluate video camera control quality arises from unknown intrinsic parameters. Similar camera motions in image frames can have different interpretations in 3D metrics; for example, a leftward motion with similar visual displacement can correspond to varying metric distances depending on the scale of the scene~\cite{zisserman2004mvg}. While some Structure-from-Motion (SfM) methods estimate camera intrinsics, these estimates are often unreliable due to limited frame numbers (often around 15) and the typically smooth nature of camera motion, with a short stereo baseline needed for accurate intrinsic estimation.

To address this issue, in our paper, we report the \textbf{scale-corrected} camera trajectory by normalizing the trajectory length to match the ground truth. Formally, this is done by rescaling the estimated trajectory \(\hat{\mathbf{T}}\) such that:
\[
\hat{\mathbf{T}}_{\text{scaled}} = \frac{\|\mathbf{T}^*\|}{\|\hat{\mathbf{T}}\|} \cdot \hat{\mathbf{T}},
\]
where \(\|\mathbf{T}^*\|\) is the length of the ground truth trajectory and \(\|\hat{\mathbf{T}}\|\) is the length of the estimated trajectory.

In our paper, all computations are performed using \texttt{evo}\footnote{\url{https://github.com/MichaelGrupp/evo}}, a standard trajectory evaluation toolbox widely used for SLAM and visual odometry evaluations.

\paragraph{Computational Efficiency}
Unfortunately, the computation time for SfM is relatively long and, most importantly, difficult to parallelize on GPU due to the sequential nature of the optimization problem. For instance, processing a single video with 16 frames using ParticleSfM~\cite{zhao2022particlesfm} can take up to 4 minutes on average, including feature extraction and the optimization pipeline required for convergence. In our study, computing SfM for 1000 generated videos required an average of 66 hours on a single CPU.

\subsection{FlowSim metric}

\paragraph{Flow Similarity.} As detailed in Section~\ref{sub:metrics} and Equation~\ref{eq:flowsim} of the main manuscript, we introduce the flow similarity metric. Similar to RPE, the concept of optical flow involves computing the relative on-image pixel motion between frames, which is widely used as an intermediate feature in SLAM and dense SfM processes~\cite{zhang2020flowfusion,cheng2019improving}. 

Compared to other on-image features, optical flow is less content-dependent and can serve as a robust metric for comparing video similarity. Unlike SfM trajectory features, optical flow does not rely on prior knowledge or precise estimation of the ground truth scaling of the scene. This makes it inherently robust to scaling ambiguities.

In our implementation, we measure the alignment of two optical flow directions while ignoring magnitudes to avoid content bias, as magnitudes are often influenced by disparity (i.e., differences in depth). Additionally, we exclude low-magnitude components because their directions are typically unreliable.

Figure~\ref{fig:supp-flow} highlights the robustness of optical flow in comparing camera motion between videos. Despite differing content, similar flows indicate similar camera motion.

\begin{figure}[htbp]
    \centering
    \begin{subfigure}{0.49\columnwidth}
        \centering
        \includegraphics[width=\linewidth]{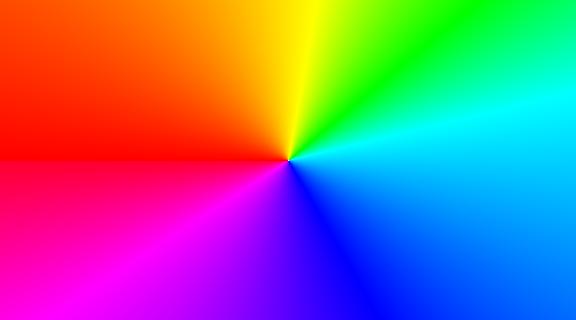}
        \caption{Theoretical flow}
        \label{supfig:zooming-flow-theoretical}
    \end{subfigure}
    \hfill
    \begin{subfigure}{0.49\columnwidth}
        \centering
        \includegraphics[width=\linewidth]{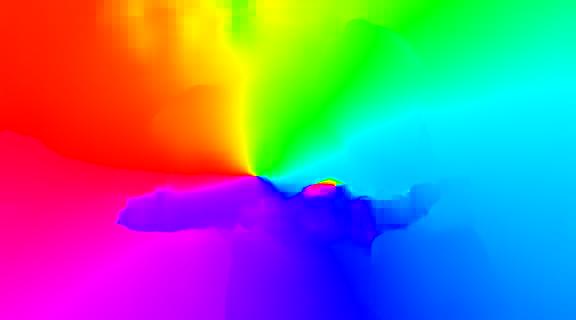}
        \caption{Generated flow}
        \label{supfig:zooming-flow-reference}
    \end{subfigure}
    \caption{\textbf{Theoretical vs. generated zooming flows.}}
    \label{supfig:zooming-flows}
\end{figure}

\paragraph{Zoom flow similarity.} To evaluate the quality of the zooming effect, we compute the flow similarity between the theoretical zooming flow and the generated one. The theoretical zooming flow is derived from Equation~\ref{eq:focal-change-image} in the main manuscript as:
\begin{equation}
    \begin{bmatrix} u_\text{fl} \\ v_\text{fl} \end{bmatrix} - \begin{bmatrix} u_{\text{fl}'} \\ v_{\text{fl}'} \end{bmatrix} = (s - s') \begin{bmatrix} u - c_x \\ v - c_y \end{bmatrix}
    \ ,
    \label{supeq:zooming-flow}
\end{equation}
where $s$ and $s'$ denote the zooming scales, and $(c_x, c_y)$ are the principal point coordinates (i.e. the screen center).  
As illustrated in Figure~\ref{supfig:zooming-flow-theoretical}, the flow direction for a zoom-in effect converges toward the principal point (or diverges outward for a zoom-out effect), aligning with Equation~\ref{supeq:zooming-flow}.
The generated flow, shown in Figure~\ref{supfig:zooming-flow-reference}, closely matches the theoretical flow, with minor deviations caused by variations in frame content.

\begin{figure}[htbp]
    \centering
    \begin{subfigure}{0.49\columnwidth}
        \centering
        \includegraphics[width=\linewidth]{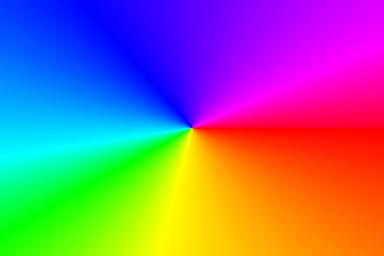}
        \caption{Theoretical flow}
        \label{supfig:distortion-flow-theoretical}
    \end{subfigure}
    \hfill
    \begin{subfigure}{0.49\columnwidth}
        \centering
        \includegraphics[width=\linewidth]{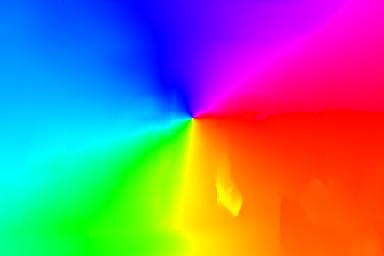}
        \caption{Generated flow}
        \label{supfig:distortion-flow-reference}
    \end{subfigure}
    \caption{\textbf{Theoretical vs. generated distorted flows.}}
    \label{supfig:distortion-flows}
\end{figure}

\paragraph{Distortion flow similarity.}
To evaluate the quality of the distortion effect, we compute the flow similarity between the theoretical distortion flow and the generated one. The theoretical distortion flow is derived from Equation~\ref{eq:distrotion} in the main manuscript as:

\begin{equation}
    \begin{bmatrix} u_\mathbf{D} \\ v_\mathbf{D} \end{bmatrix} - \begin{bmatrix} u_{\mathbf{D}'} \\ v_{\mathbf{D}'} \end{bmatrix} = \begin{bmatrix} u \\ v \end{bmatrix} (\mathbf{D} -  \mathbf{D}') \begin{bmatrix} r^2 \\ r^4 \\ r^6 \end{bmatrix}
    \ ,
    \label{supeq:distortion-flow}
\end{equation}

where $\mathbf{D}$ and $\mathbf{D}'$ denote the distortion parameters and $r= \sqrt{(u-c_x)^2+(v-c_y)^2}$ distance of each pixel towards image center.  
As illustrated in Figure~\ref{supfig:distortion-flow-theoretical}, the flow direction for a distortion effect also converges toward the principal point —or diverges outward depending on the sign of $\mathbf{D}-\mathbf{D}'$—, aligning with Equation~\ref{supeq:distortion-flow}.
The generated flow, shown in Figure~\ref{supfig:distortion-flow-reference}, closely matches the theoretical flow, again, with minor deviations caused by variations in frame content.

\paragraph{Computational Efficiency}
During implementation, we use RAFT~\cite{teed2020raft}, a fast deep optical flow estimator with GPU-based implementation, which significantly speeds up the flow estimation process compared to SfM and can be easily parallelized on GPU. For example, computing optical flows for 1000 generated videos takes approximately 10 minutes on a GPU, compared to 66 hours required by SfM on a single CPU.

\paragraph{As a result, our key messages concerning the evaluation metrics are:}
\begin{enumerate}
    \item Pose metrics for assessing camera control in generated videos are intrinsically less accurate, particularly when directly computing APE. 
    \item Using RPE with scale correction techniques improves robustness; however, the computational cost is prohibitively high for scaling to AI-generated videos.
    \item We propose optical flow similarity (FlowSim), based on flow direction, offers a viable alternative. It serves as a good approximation of RPE while being computationally efficient, fast to compute, and scalable for large-scale AI-generated videos.
    \item The flow similarity metric can also be used to confirm other optic features than motion, such as focal length change (zoom) and distortions.
\end{enumerate}

\section{Additional qualitative results}
\label{supsec:quali}
We provide additional qualitative results in Figure~\ref{fig:supp-quals}, demonstrating that AKiRa method performs well in several key aspects: accurately capturing camera motion (trees) with high video quality (the middle stormtrooper’s head cf. CameraCtrl); maintaining consistency during zooming (mountain peak, cat's frame); effectively reflecting distortion effects (gift, bird’s-eye view of Barcelona city); and rendering various aperture and bokeh effects (surfing on the beach, grass) with controllability.


%% file: tab/supp-sota.tex

{{\renewcommand{\arraystretch}{1} 
\begin{table*}[t]
\centering
\resizebox{\textwidth}{!}{
\begin{tabular}{cl|cc|ccc|ccc}
\toprule
\multicolumn{2}{c|}{\textbf{Method}} & \multicolumn{2}{c|}{\textbf{Video quality}} & \multicolumn{3}{c|}{\textbf{Camera motion fidelity}} & \multicolumn{3}{c}{\textbf{Dynamic consistency (VBench)}} \\ 
Backbone & Camera control & FVD $\downarrow$ & CD-FVD $\downarrow$ & RPE-R (deg)$\downarrow$ &  RPE-t (cm)$\downarrow$ & FlowSim $\uparrow$ &  Consistency $\uparrow$  & Smoothness $\uparrow$ & Flickering$\uparrow$  \\ 
\midrule
\multirow{3}{*}{\rotatebox[origin=c]{0}{\shortstack{AnimateDiff\\\cite{guo2023animatediff}}}}
& MotionCtrl~\cite{wang2024motionctrl}  & $237.22$ & $543.24$ & $0.387$ & $1.536$ & $67.83$ & $0.9779$ & $0.9834$ & $0.9712$ \\ 
& CameraCtrl~\cite{he2024cameractrl}    & $177.70$ & $106.24$ & $0.377$ & $1.555$ & $77.08$ & $0.9779$ & $0.9834$ & $0.9712$ \\ 
& AKiRa (ours)                          & \cellcolor{tabfirst}$128.55$ & \cellcolor{tabfirst}$89.16$  & \cellcolor{tabfirst}$0.323$ & \cellcolor{tabfirst}$1.347$ & \cellcolor{tabfirst}$84.04$ &  \cellcolor{tabfirst}$0.9809$ & \cellcolor{tabfirst}$0.9882$ & \cellcolor{tabfirst}$0.9745$ \\ 
\cline{1-10}
\multirow{3}{*}{\rotatebox[origin=c]{0}{\shortstack{SVD\\\cite{blattmann2023svd}}}} 
& MotionCtrl~\cite{wang2024motionctrl} & $122.67$ & $330.75$ & $1.030$ & $1.326$ & $24.14$ & $0.9516$ & $0.9814$ & $0.9404$ \\
& CameraCtrl~\cite{he2024cameractrl}   & $55.55$  & $50.18$  & \cellcolor{tabfirst}$0.312$ & $1.268$ & \cellcolor{tabfirst}$92.19$  & $0.9836$ & $0.9928$ & $0.9695$ \\ 
& AKiRa (ours)                         & \cellcolor{tabfirst}$54.83$  & \cellcolor{tabfirst}$41.55$  & \cellcolor{tabfirst}$0.312$ & \cellcolor{tabfirst}$1.236$ & $91.51$  & \cellcolor{tabfirst}$0.9851$ & \cellcolor{tabfirst}$0.9933$ & \cellcolor{tabfirst}$0.9733$ \\
\bottomrule
\end{tabular}}
\caption{\textbf{Comparison with the state-of-the-art.} 
Comparison of AKiRa and concurrent methods with different backbones on RealEstate dataset, evaluating video quality, camera motion fidelity, and dynamic consistency. 
\colorbox{tabfirst}{Best}.}
\label{supptab:sota}
\end{table*}}}

%% file: fig/supmat_bokeh.tex
\begin{figure*}[htbp]
    \centering
    \begin{subfigure}{0.49\textwidth}
        \centering
        \includegraphics[width=\linewidth]{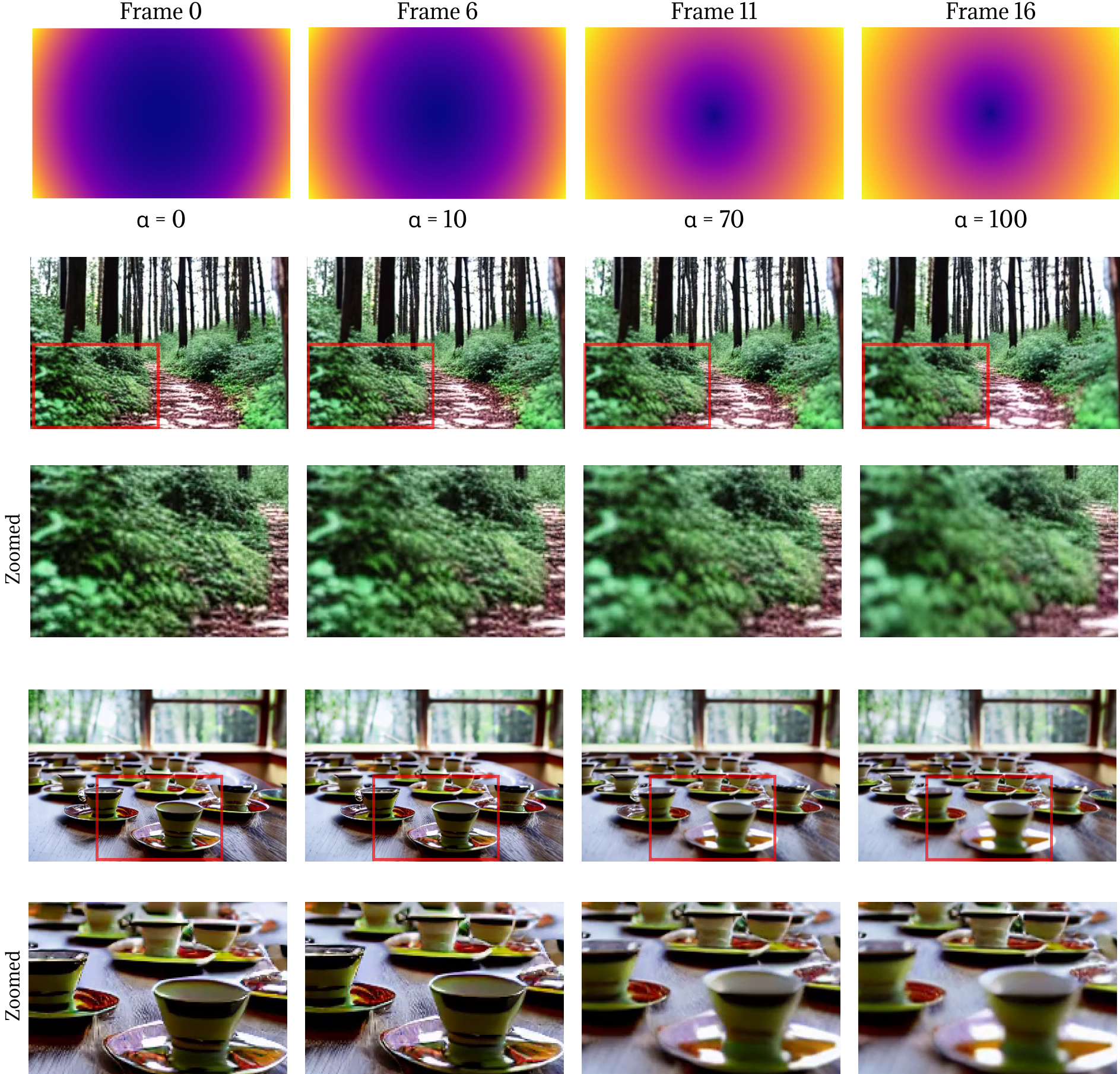}
        \caption{Example of increasing apertures along video time}
        \label{fig:qual-bokeh-aperture}
    \end{subfigure}
    \hfill
    \begin{subfigure}{0.49\textwidth}
        \centering
        \includegraphics[width=\linewidth]{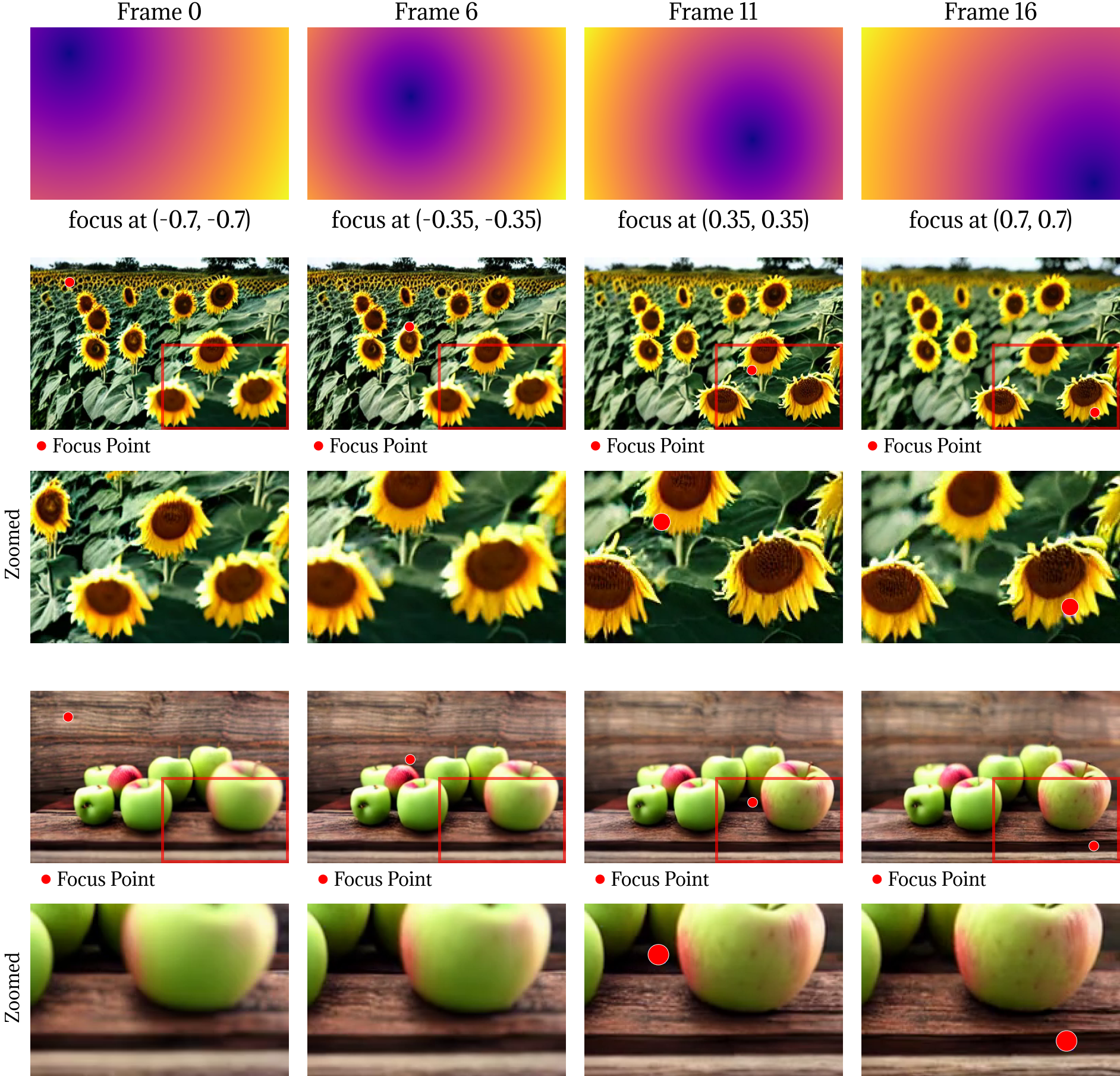}
        \caption{Example of shifting focus point (red dot) along video time}
        \label{fig:qual-bokeh-focus-point}
    \end{subfigure}
    \caption{\textbf{Qualitative Results.} We demonstrate the qualitative performance of \akira on bokeh variations for both (a) aperture levels and (b) focus points. The results are generated using Animatediff~\cite{guo2023animatediff}. In (a), we observe that the blur area and intensity increase proportionally with the aperture parameter \(\alpha\). In (b), the blur area shifts dynamically following changes in the focus point, showing the effectiveness of \akira in handling variations in depth of field.}
    \label{fig:sub-quals}
\end{figure*}

%% file: tab/supp-aperture.tex
\begin{table}[t]
\centering
\resizebox{\columnwidth}{!}{
\begin{tabular}{c|cc|c|ccc}
\toprule
\multirow{2}{*}{\textbf{Apert.}} & \multicolumn{2}{c|}{\textbf{Video quality}} & \multicolumn{1}{c|}{\textbf{Motion fidelity}} & \multicolumn{3}{c}{\textbf{Dynamic consistency (VBench)}} \\ 
 & FVD $\downarrow$ & CD-FVD $\downarrow$ & FlowSim $\uparrow$ &  Consistency $\uparrow$  & Smoothness $\uparrow$ & Flickering$\uparrow$  \\ 
\midrule
0   & $350.05$ & $333.61$ & $70.82$ & \cellcolor{tabfirst}$0.9697$ & $0.9702$ & $0.9525$ \\ 
5   & $353.09$ & $337.48$ & $70.94$ & $0.9695$ & $0.9705$ & $0.9528$ \\ 
10  & $354.94$ & $334.95$ & $71.15$ & $0.9692$ & $0.9709$ & $0.9534$ \\ 
30  & $341.31$ & \cellcolor{tabfirst}$327.02$ & \cellcolor{tabfirst}$71.23$ & $0.9686$ & $0.9728$ & $0.9559$ \\ 
50  & \cellcolor{tabfirst}$332.27$ & $328.74$ & $70.97$ & $0.9686$ & $0.9735$ & $0.9572$ \\ 
100 & $342.77$ & $328.13$ & $70.82$ & $0.9683$ & \cellcolor{tabfirst}$0.9737$ & \cellcolor{tabfirst}$0.9574$ \\ 
\bottomrule
\end{tabular}}
\caption{\textbf{Influence of aperture.} 
Influence of aperture effect on AKiRa's performances with Animatediff backbone on RealEstate dataset. 
\colorbox{tabfirst}{Best}.}
\label{supptab:aperture}
\end{table}

%% file: fig/supp-quals.tex
\begin{figure*}[htbp!]
    \centering
    \begin{subfigure}{0.48\textwidth}
        \centering
        \includegraphics[width=\linewidth]{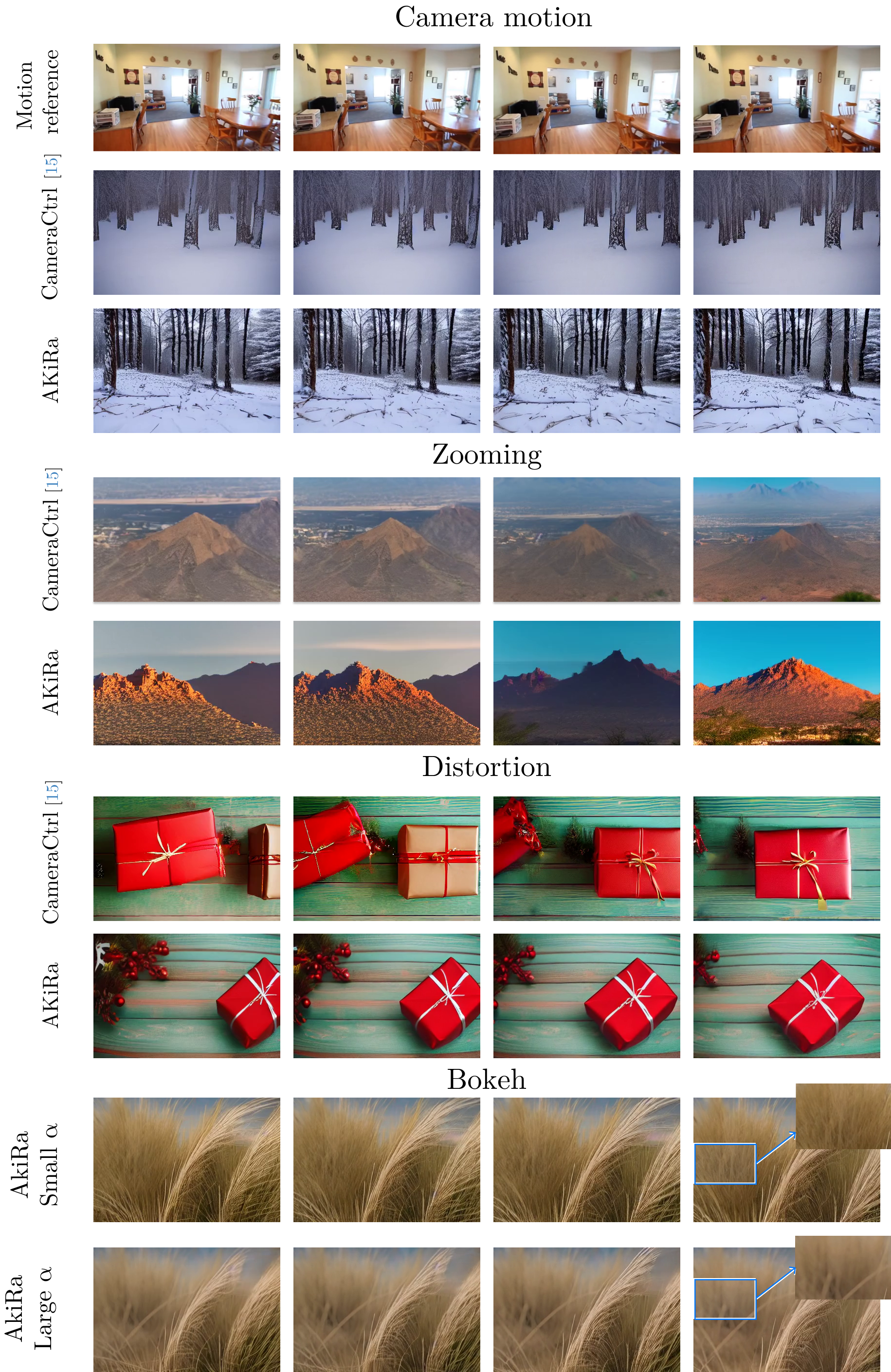}
        \caption{Animatediff~\cite{guo2023animatediff}}
    \end{subfigure}
    \hfill
    \begin{subfigure}{0.505\textwidth}
        \centering
        \includegraphics[width=\linewidth]{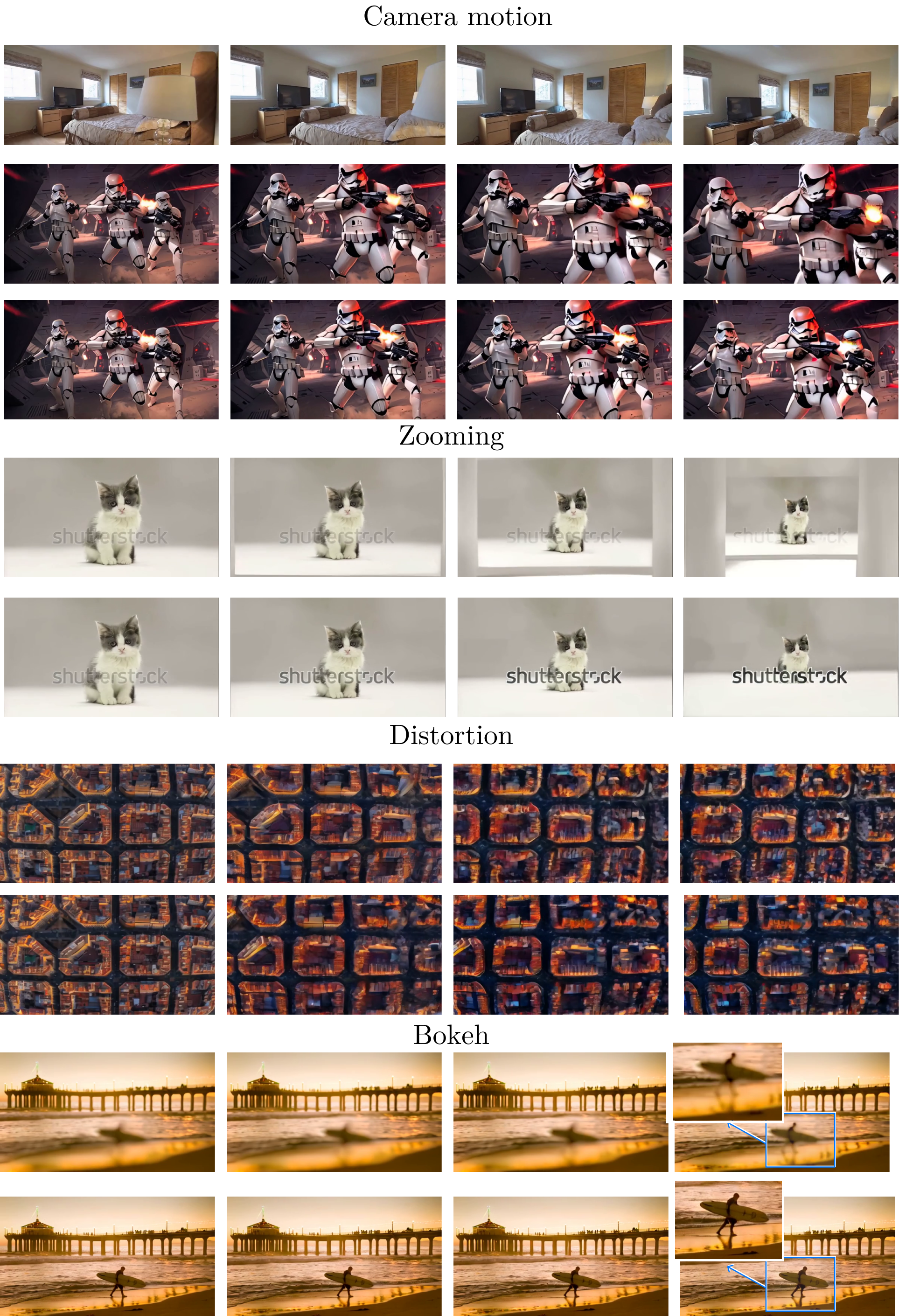}
        \caption{SVD~\cite{blattmann2023svd}}
    \end{subfigure}
    \caption{\textbf{Qualitative results} of \akira on Animatediff~\cite{guo2023animatediff} and SVD~\cite{blattmann2023svd} backbones. We recommend viewing the supplementary video.}
    \label{fig:supp-quals}
\end{figure*}